%% file: main.tex
\PassOptionsToPackage{hypertexnames=false}{hyperref}
\documentclass[pdflatex,sn-mathphys-ay]{sn-jnl}

\usepackage{graphicx}%
\usepackage{multirow}%
\usepackage{amsmath,amssymb,amsfonts}%
\usepackage{amsthm}%
\usepackage[title]{appendix}%
\usepackage{xcolor}%
\usepackage{textcomp}%
\usepackage{manyfoot}%
\usepackage{booktabs}%
\usepackage{algorithm}%
\usepackage{algorithmicx}%
\usepackage{algpseudocode}%
\usepackage{listings}%
\usepackage{threeparttable}
\usepackage{array}
\usepackage{tabularx}
\usepackage{microtype}

\theoremstyle{thmstyleone}%
\theoremstyle{thmstyletwo}%
\theoremstyle{thmstylethree}%
\usepackage[inline]{enumitem}
\begin{document}

\title[osmAG-Nav]{osmAG-Nav: A Hierarchical Semantic Topometric Navigation Stack for Robust Lifelong Indoor Autonomy}


\author[1]{\sur{Yongqi Zhang}}
\equalcont{These authors contributed equally to this work.}

\author[1]{\sur{Jiajie Zhang}}
\equalcont{These authors contributed equally to this work.}

\author[1]{\sur{Chengqian Li}}

\author[1]{\sur{Fujing Xie}}

\author*[1]{\sur{S\"oren Schwertfeger}}\email{soerensch@shanghaitech.edu.cn}

\affil*[1]{\orgdiv{School of Computer Science and Technology}, \orgname{ShanghaiTech University}, \orgaddress{\street{393 Huaxia Road}, \city{Shanghai}, \postcode{201210}, \state{Shanghai}, \country{China}}}


\abstract{The deployment of mobile robots in large-scale, multi-floor environments demands navigation systems that achieve \textit{spatial scalability} without compromising local kinematic precision. Traditional navigation stacks, predominantly reliant on monolithic occupancy grid maps, face severe bottlenecks in storage efficiency, cross-floor reasoning, and long-horizon planning. To address these systemic limitations, this paper presents \textbf{osmAG-Nav}, a complete, open-source navigation stack for ROS2 built upon the hierarchical semantic topometric \textbf{OpenStreetMap Area Graph (osmAG)} map standard. The system follows a ``System of Systems'' architecture that decouples global topological reasoning from local metric execution. A Hierarchical osmAG planner replaces dense grid searches with an LCA-anchored planning pipeline on a \textit{passage-centric graph} whose edge costs are derived from local raster traversability rather than Euclidean distance, yielding low-millisecond planning even on long campus-scale routes. A \textit{Rolling Window} mechanism rasterizes only a fixed-size local metric grid around the robot, keeping the local costmap memory footprint independent of the total mapped area, while a \textit{Segmented Execution} strategy dispatches intermediate goals sequentially to standard ROS2 controllers for smooth, continuous handoffs. System robustness is reinforced by a structure-aware LiDAR localization framework that filters dynamic clutter against permanent architectural priors. Extensive experiments on a real-world multi-story indoor-outdoor campus ($>$11,025\,m$^2$) show that, on the comparable same-floor benchmark subset, osmAG-Nav delivers up to $7816\times$ lower planning latency than a grid-based baseline on long routes while maintaining low path-length overhead and lifelong localization stability. A single-floor long-range robot mission further validates integrated stack reliability. The full stack is released as modular ROS2 Lifecycle Nodes.}





\maketitle

\input{sections/1_intro}

\input{sections/2_related_work}

\input{sections/3_osmAG_standard}

\input{sections/4_sys_architecture}

\input{sections/5_navigation}

\input{sections/6_localization}

\input{sections/7_sys_integration}

\input{sections/8_experiment}

\input{sections/9_conclusion}

\backmatter

\input{sections/10_declarations}

\clearpage

\begin{appendices}

\input{sections/appendix_osmag_reference}

\end{appendices}

\clearpage

\bigskip






\bibliography{sn-bibliography}

\end{document}

%% file: sections/1_intro.tex
\section{Introduction}
\label{sec:intro}

In recent years, the operational domain of autonomous mobile robots has expanded significantly, evolving from confined laboratory settings to complex, large-scale, and dynamic environments such as hospitals, office campuses, and warehouses~\citep{hawes2017strands}. As the demand for long-term autonomy increases, the navigation systems underpinning these robots must demonstrate not only robustness against environmental changes but also scalability across multi-floor infrastructures. Early pioneering works like RHINO~\citep{burgard1999experiences} and Minerva~\citep{thrun1999minerva} demonstrated the feasibility of deploying robots in public spaces for days or weeks. Building on this legacy, the Robot Operating System (ROS) introduced \texttt{move\_base}~\citep{marder2010office}, a modular navigation framework that became the \textit{de facto} standard for mobile robot deployment in research and industry. More recently, ROS2~\citep{maruyama2016exploring} and its Navigation Stack (Nav2)~\citep{macenski2020marathon} have further standardized the architecture with improved real-time performance and lifecycle management. However, despite these advancements, the underlying reliance on dense metric representations---predominantly 2D occupancy grid maps~\citep{thrun2002probabilistic}---remains a persistent barrier to scalability.

\begin{figure}[t]
    \centering
    \includegraphics[width=\linewidth]{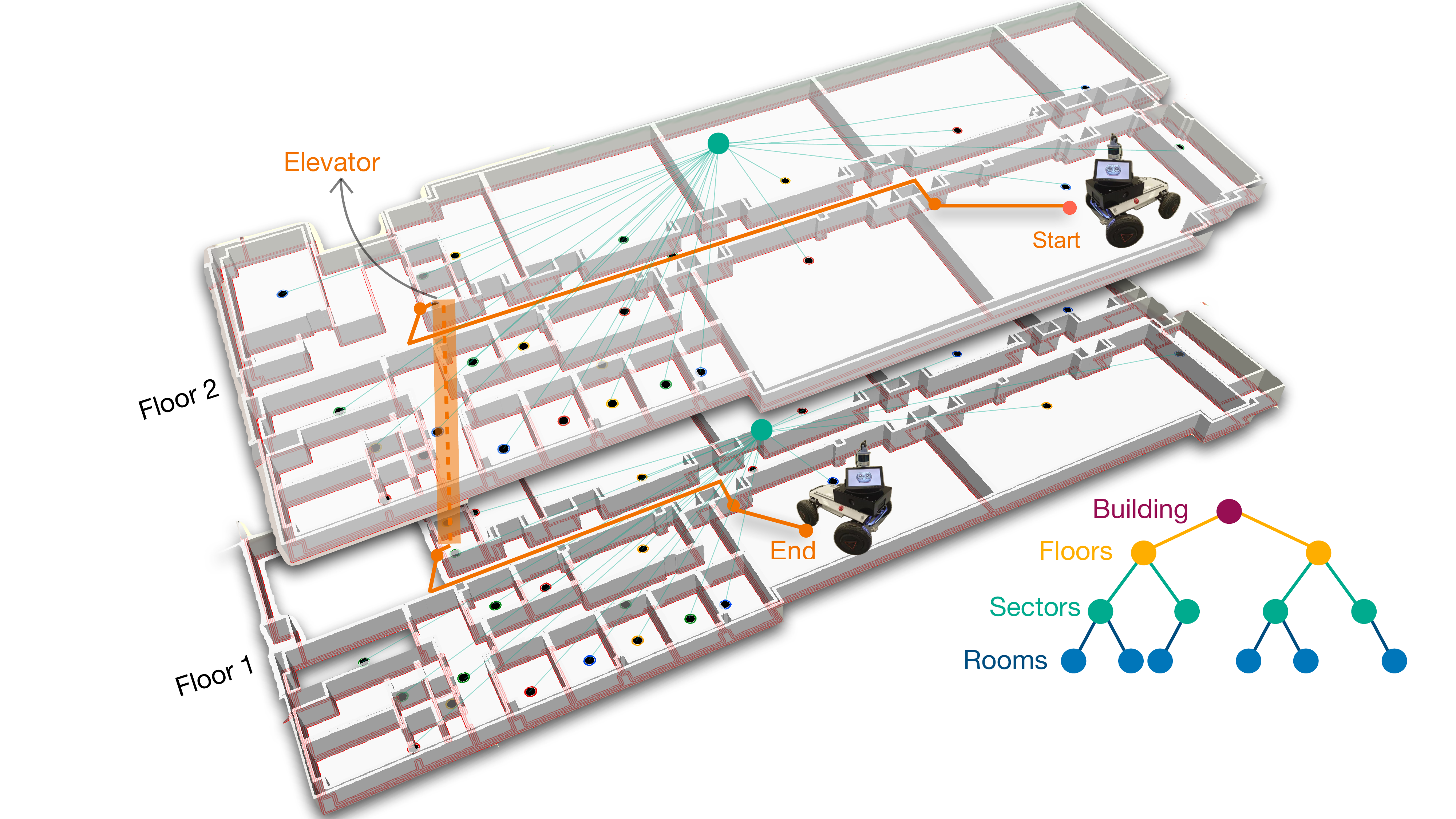} 
    \caption{\textbf{Hierarchical Multi-Floor Navigation with osmAG-Nav.} 
    The proposed system leverages the \textit{osmAG} standard to abstract complex environments into a semantic hierarchy (bottom right tree), where atomic locations (Rooms, blue nodes) are logically grouped into Sectors (teal nodes). 
    This structural abstraction enables efficient global routing across large-scale facilities. 
    The visualization demonstrates an autonomous cross-floor mission: the planner generates a topology-centric path (orange trajectory) that navigates from a 2nd-floor Start pose, identifies and utilizes the elevator (highlighted vertical shaft) as a semantic passage, and seamlessly proceeds to the 1st-floor End goal. 
    By defining the environment through permanent vector structures (walls) and topological connections rather than dense metric grids, the system maintains scalability while keeping the runtime local metric map memory footprint bounded independently of the total mapped area.}
    \label{fig:teaser}
\end{figure}

While occupancy grids have served the community well for decades, their limitations become acute when scaling to lifelong, multi-floor operations. Even with efficient data structures such as octrees~\citep{hornung2013octomap}, the sheer volume of metric data in campus-scale environments imposes heavy memory and computational burdens, often forcing a trade-off between global coverage and local agility~\citep{thrun1998learning}. Beyond storage constraints, grid maps are inherently geometric; they lack the high-level semantic understanding---distinguishing a ``corridor'' from an ``elevator''---that is essential for complex task execution and human-robot interaction~\citep{kostavelis2015semantic}. Furthermore, maps constructed via traditional Simultaneous Localization and Mapping (SLAM) often capture transient features like furniture. This rigidity leads to map obsolescence and localization instability as the environment inevitably changes~\citep{macenski2020marathon, krajnik2016frequency}. Perhaps most critically, standard 2D grids struggle to natively represent the vertical topology of modern infrastructure, often necessitating fragile, ad-hoc state machines to manage elevator transitions~\citep{palacin2023procedure}. These compounding limitations demand a fundamental rethinking of the navigation paradigm itself.

Rather than pursuing incremental improvements to the grid-based workflow, we propose a paradigm shift from geometry-centric to \textit{topology-centric} navigation. We argue that for persistent operation in man-made environments, robots should leverage the inherent structural hierarchy of buildings rather than relying solely on raw sensory snapshots. Building upon the \textit{osmAG} (OpenStreetMap Area Graph) standard introduced in our prior work~\citep{feng2023osmag}, this paper presents \textbf{osmAG-Nav}, a complete, open-source navigation stack for ROS2 that utilizes hierarchical, semantic, topometric maps. Complementing this runtime stack, we leverage an automated pipeline to generate maps directly from architectural CAD files~\citep{zhang2025generation}, thereby streamlining deployment and solving the ``cold start'' mapping problem.

The practical viability of this topology-centric approach hinges on a \textbf{hierarchical planning architecture} that restructures both the map representation and the planning pipeline. Unlike conventional systems that perform A* search over high-resolution global costmaps, our planner operates on a lightweight \textit{passage-centric graph} whose leaf-area edge costs are derived from local raster traversability rather than straight-line geometry, reducing query times by orders of magnitude compared to grid-based search~\citep{hart1968formal} while making route selection less sensitive to misleading indoor shortcuts. To bridge abstract topological routes with the robot's kinematic requirements, a \textit{Rolling Window} mechanism rasterizes only a fixed-size local metric window around the robot, keeping the local costmap memory footprint bounded independently of the total environment size. A \textit{Segmented Execution} strategy then dispatches intermediate goals sequentially via the Nav2 action interface, handling cross-floor transitions without monolithic path computation. This design draws on foundational work in spatial semantic hierarchies~\citep{kuipers2000spatial} and hybrid metric-topological navigation~\citep{konolige2011navigation}. To ensure long-term operational stability, the system integrates a robust localization module adapted from our prior work~\citep{xie2023robust}, which filters dynamic clutter by matching LiDAR data exclusively against permanent structural features and handles geometric degeneracy through direction-aware constraint selection.


The major contributions of this work are:

\begin{itemize}
    \item \textbf{Formalization of the osmAG Standard for Robotics \& Embodied AI:} 
    We formally define and open-source the \textit{OpenStreetMap Area Graph (osmAG)} standard. Going beyond a task-specific navigation format, osmAG provides a general-purpose hierarchical semantic spatial representation. Its applicability has been demonstrated in WiFi localization~\citep{ma2025wifi}, LLM-assisted path reasoning~\citep{xie2025intelligent}, semantic knowledge mapping~\citep{xie2024empowering}, and object-goal navigation~\citep{xie2026osmag}, indicating its utility as a shared spatial backbone across robotics and embodied AI tasks.

    \item \textbf{A Decoupled ``System of Systems'' Architecture:} 
    We propose an architectural paradigm that decouples global topological reasoning from local metric execution. Unlike monolithic architectures that bind planning to dense global grids, our approach positions the osmAG stack as a high-level executive that orchestrates a standard Nav2 local stack. This ``System of Systems'' design bridges the gap between abstract topological routing and precise kinematic control.

    \item \textbf{The osmAG-Nav Stack---Core Technical Capabilities:} 
    We present a complete, open-source navigation stack that overcomes the systemic limitations of traditional grid-based systems:

    \begin{enumerate}[label=(\roman*)]
        \item \textit{Rolling Window Rasterization:} We introduce a fixed-size \textit{Rolling Window} mechanism that renders local grid maps on demand from the vector map, keeping the runtime local costmap memory footprint bounded independently of the total environment size;
         \item \textit{Native Multi-Level Topology:} Explicit modeling of vertical connectivity (e.g., elevators) enables coherent cross-floor reasoning without planar constraints.
    \end{enumerate}

    \item \textbf{Hierarchical osmAG Planning Engine:} 
    We propose a hierarchical osmAG planner that exploits the map's hierarchical topometric structure through an LCA-based query decomposition. Operating on a \textit{passage-centric graph} with \textit{Segmented Execution}, it specifically addresses the computational scalability challenge of long-horizon routing. On the common same-floor comparable subset of our campus benchmark, Grid A* rises from $204.3$ ms on short queries to $10641.0$ ms on long queries, whereas the hierarchical planner remains in the low-millisecond regime and achieves $1995\times$ and $7816\times$ speedups over Grid A* on the medium and long buckets, respectively, while incurring only a small constant overhead on the shortest routes.

    \item \textbf{Improved Robust LiDAR Localization:} 
    Building upon our prior work \citep{xie2023robust}, we integrate a structure-aware LiDAR localization framework tailored for long-term ROS2 deployment. Key algorithmic improvements include a probabilistic odometry fusion module with adaptive weighting to bridge LiDAR dropouts, optimization divergence protection via dynamic step-limiting, and direction-aware geometric constraints that selectively preserve orthogonal features to mitigate aliasing in featureless corridors. These mechanisms collectively strengthen tracking stability in dynamic and geometrically degenerate environments.

    \item \textbf{Campus-Scale Real-World Validation:} 
    We conduct extensive experiments on a real-world multi-story indoor-outdoor campus environment ($>11,025\,m^2$). Offline benchmarks rigorously validate the computational efficiency of the hierarchical planner and the increased robustness of the localization module. A single-floor long-range mission covering approximately 235\,m further demonstrates the reliability of the integrated navigation stack for autonomous long-distance operation.
\end{itemize}


The remainder of this paper is organized as follows: Section~\ref{sec:related_work} reviews related work on map representations, localization, planning, and navigation frameworks. Section~\ref{sec:osmag} formalizes the osmAG standard and its hierarchical semantic-topometric representation. Section~\ref{sec:sys_architecture} presents the overall system architecture and ROS2 design philosophy. Section~\ref{sec:nav} details the hierarchical navigation methodology, including rolling-window map serving, physically-aware raster-cost graph construction, LCA-based search, and segmented execution. Section~\ref{sec:loc} describes the robust localization framework. Section~\ref{sec:system_integration} summarizes the automated map-generation pipeline and software integration. Section~\ref{sec:experiments} reports the experimental evaluation, and Section~\ref{sec:conclusion} concludes the paper.

%% file: sections/2_related_work.tex
\section{Related Work}
\label{sec:related_work}

Research most relevant to osmAG-Nav falls into four strands: map representation, indoor localization, large-scale planning, and navigation software frameworks. Early systems such as RHINO~\citep{burgard1999experiences} and Minerva~\citep{thrun1999minerva} established the viability of long-term autonomy in public spaces with probabilistic localization and grid-based maps. More recent work has improved each strand substantially, yet the combination of semantic indoor structure, multi-floor reasoning, and ROS2-native deployment remains underdeveloped. We review that gap below.

\subsection{Map Representations}
\label{subsec:map_representations}
Map representation determines which planning and localization strategies remain practical at scale. Conventional mobile robot navigation has been dominated by \textit{metric maps}, especially 2D occupancy grids~\citep{elfes1989using}. Layered costmaps~\citep{lu2014layered} improve flexibility by combining static structure, sensed obstacles, and inflation policies, but they remain tied to discrete cells. As environment size and floor count increase, both memory use and global search cost increase accordingly.

\textit{Topological maps} address this scalability issue by representing connectivity explicitly. Early systems such as DERVISH~\citep{nourbakhsh1995dervish} demonstrated the practical value of topological state representations, while later \textit{hybrid metric-topological} systems combined sparse global structure with local metric execution~\citep{thrun1998learning, konolige2011navigation}. Unlike purely metric maps, these representations support efficient long-range reasoning, but they often lack the geometric detail needed for precise local motion.

Recent work has pushed this direction toward \textit{semantic} and \textit{hierarchical} models. The Spatial Semantic Hierarchy~\citep{kuipers2000spatial} formalized layered spatial knowledge, and richer 3D semantic representations---including scene graphs~\citep{armeni20193d} and metric-semantic mapping systems such as Kimera~\citep{rosinol2020kimera}---extend this idea to perception-rich 3D scenes. Their expressive power is attractive, but it usually comes with substantial storage and computation overhead due to dense geometry such as point clouds or meshes.

Vector standards have proven especially effective in autonomous driving. Lanelet2~\citep{poggenhans2018lanelet2}, built on the OpenStreetMap (OSM) data model, provides an efficient vector-topological representation for high-definition road maps. For indoor environments, standards such as OGC IndoorGML~\citep{li2016indoorgml} offer a formal schema for spatial modeling, but translating such GIS-oriented models into real-time robot navigation still requires considerable engineering effort. Our work builds on osmAG~\citep{feng2023osmag}, which provides a lightweight hierarchical semantic-topological vector representation tailored to rooms, corridors, passages, and multi-floor indoor structure.

\subsection{Indoor Localization}
\label{subsec:indoor_localization}
SLAM systems such as Cartographer~\citep{hess2016real} and LIO-SAM~\citep{shan2020lio} remain central to initial map construction and odometry estimation. For long-term deployment, however, many systems localize against a prior map instead of continuously extending it, partly to avoid drift accumulation and map inconsistency. In this setting, odometric uncertainty is often moderated through sensor fusion frameworks such as EKF-based estimators~\citep{moore2015generalized}, while global consistency still depends on map matching.

Within indoor robotics, occupancy-grid localization remains the dominant baseline. AMCL and related scan-to-map particle-filter approaches compare incoming scans against a rasterized prior map, while adaptive sampling mechanisms such as KLD-sampling~\citep{fox2001kld} control particle count according to uncertainty. In contrast to structure-based localization, occupancy-grid matching is highly sensitive to long-term scene change: moved furniture, temporary obstacles, and human traffic alter scan appearance relative to the static map and can degrade localization severely.

An alternative direction is to localize against lightweight vector priors rather than dense metric reconstructions. \citet{floros2013openstreetslam} anchored visual odometry to OpenStreetMap road networks for global vehicle localization, and \citet{ruchti2015localization} extended this idea to 3D LiDAR localization against OSM-derived road structure. These studies show that stable structural priors can support localization without requiring a dense pre-built metric map. Our localization stack follows this line in the indoor domain by building on AGLoc~\citep{xie2023robust}, which aligns sensor observations to permanent architectural elements instead of transient clutter.

\subsection{Path Planning}
\label{subsec:path_planning}
Path planning in large buildings must balance route quality against computational tractability. A globally optimal search on a high-resolution multi-floor grid is often impractical, which motivates sparse abstractions and hierarchical decomposition. Sampling-based roadmaps such as PRM~\citep{kavraki1996probabilistic} introduced the idea of precomputing sparse connectivity, while the Spatial Semantic Hierarchy~\citep{kuipers2000spatial} showed how topological abstraction can organize long-range planning at multiple semantic scales. Hybrid systems~\citep{konolige2011navigation} then made this split operational by planning globally on sparse structure and executing locally in metric space.

Multi-floor navigation makes the problem harder because robots must reason about elevators and stairs as well as horizontal passage. \citet{palacin2023procedure} highlights the procedural complexity of elevator interaction. Unlike approaches that rely on external state machines for floor switching, graph-based formulations can treat vertical transitions as part of the planning problem itself.

For local collision avoidance and short-horizon motion generation, controllers such as DWA~\citep{fox2002dynamic} and TEB~\citep{rosmann2015timed} remain widely used. These methods are effective within a local costmap, but they do not directly encode building-scale semantic constraints. Our work follows the hybrid planning tradition, but performs global reasoning on a hierarchical passage-centric graph defined by osmAG and reserves local geometric handling for the controller in a bounded neighborhood.

\subsection{Navigation Frameworks}
\label{subsec:navigation_systems}
Before standardized middleware became widespread, navigation systems were usually engineered as custom, tightly integrated stacks for specific deployments. DERVISH~\citep{nourbakhsh1995dervish} demonstrated reliable office navigation with topological state representations, while RHINO~\citep{burgard1999experiences} and MINERVA~\citep{thrun1999minerva} showed that probabilistic localization and planning could support robust public-space autonomy. For outdoor settings, \citet{kummerle2013navigation} reported a navigation system capable of operating in crowded urban environments while accounting for dynamic obstacles and traversability.

ROS shifted the field toward modular software composition by separating hardware abstraction from reusable algorithmic components. Building on ROS, the navigation stack centered around \texttt{move\_base}~\citep{marder2010office} established a widely adopted architectural pattern in which global planning, local control, and sensing are separated into cooperating modules. Nav2~\citep{macenski2020marathon} extends this model into ROS2 and further strengthens it through Behavior Trees~\citep{ogren2022behavior} and Lifecycle Nodes, improving reliability and deployment readiness.

The limitation is not the software architecture itself, but the default data model underneath it. Mainstream stacks still assume an occupancy-grid-centric world model, so semantic reasoning, native multi-floor topology, and large-scale persistence are not first-class capabilities. Our work therefore preserves the robust software architecture of Nav2 while replacing the global representation and deliberative reasoning layer with an osmAG-based semantic-topological stack.

%% file: sections/3_osmAG_standard.tex
\section{The osmAG Standard}
\label{sec:osmag}

The \textit{OpenStreetMap Area Graph} (osmAG) was introduced in prior work as a hierarchical semantic-topometric representation for mobile robotics~\citep{feng2023osmag} and has since been used for structure-based LiDAR localization~\citep{xie2023robust}, WiFi-based global localization with structural priors~\citep{ma2025wifi}, LLM-assisted path reasoning over map topology and hierarchy~\citep{xie2024empowering}, intelligent navigation with external information~\citep{xie2025intelligent}, and object-goal navigation through semantic enrichment~\citep{xie2026osmag}. However, those papers employed osmAG in task-specific forms and did not consolidate it into a single official specification. This section therefore formalizes osmAG as a unified standard by defining a normative core map specification together with sanctioned extension profiles for navigation semantics and Embodied AI semantics.

\subsection{Design Goals and Standard Boundary}
\label{subsec:osmag_scope}

osmAG is intended as a robotics and embodied-AI spatial standard rather than merely a file serialization. Its design goals are fourfold. First, it preserves direct compatibility with the OpenStreetMap (OSM) XML ecosystem so that existing authoring and visualization tools remain usable. Second, it represents environments sparsely through persistent vector structures rather than dense global grids, making the representation suitable for large buildings and campuses. Third, it natively models hierarchy, semantics, and topometric anchoring in one map. Fourth, it treats vertical topology---especially elevators and stairs---as first-class spatial structure rather than as planner-side special cases.

For clarity, the specification is organized in two layers. The \emph{core map standard} (Sections~\ref{subsec:osmag_core_model}--\ref{subsec:osmag_vertical}) defines persistent spatial primitives, hierarchy, validity constraints, and OSM/XML serialization rules. The \emph{extension profiles} (Section~\ref{subsec:osmag_extensions}) add sanctioned task-specific semantics on top of the same geometric and hierarchical backbone.

\subsection{Formal Core Model}
\label{subsec:osmag_core_model}

The canonical osmAG map is an undirected \emph{Area Graph}
\begin{equation}
    \mathcal{G}_{\mathrm{osmAG}} = (\mathcal{A}, \mathcal{P}),
\end{equation}
where each $A \in \mathcal{A}$ is an \textbf{Area} and each $P \in \mathcal{P}$ is a \textbf{Passage} connecting an unordered pair of adjacent areas. Conceptually, areas are the primary graph vertices and passages are the graph edges. On disk, both are serialized as OSM \texttt{way} elements so that the topology and its associated geometry remain explicitly editable and visualizable.

\textbf{Area.} An area is a closed polygon defined by an ordered list of OSM nodes. Areas capture navigable spaces and spatial containers at multiple scales, from rooms and corridors to floors and whole buildings. We distinguish two canonical categories:
\begin{itemize}
    \item \textbf{Inner areas:} traversable or directly localizable spaces such as rooms, corridors, elevators, and stairs;
    \item \textbf{Structure areas:} container regions that organize child areas hierarchically, such as floors, wings, sectors, or buildings.
\end{itemize}

\textbf{Passage.} A passage is the explicit topological connector between two areas. Geometrically, it is represented as a line segment or short polyline embedded on the shared boundary or interface between the connected areas. Typical examples include doors, corridor openings, and vertical transfer links.

\textbf{Root node and topometric anchoring.} Every osmAG map contains a designated \texttt{root} node that anchors the local Cartesian frame to geodetic coordinates. This node provides the reference needed to convert between OSM latitude/longitude storage and the local Cartesian frame used by robotic planning, localization, and visualization.

\textbf{Hierarchy.} Non-top-level areas may reference exactly one parent area, thereby forming a tree of containment relations such as building $\supset$ floor $\supset$ sector $\supset$ room. This hierarchy supports both logical aggregation and multi-level stacking.

\begin{figure*}[t]
    \centering
    \includegraphics[width=0.98\textwidth]{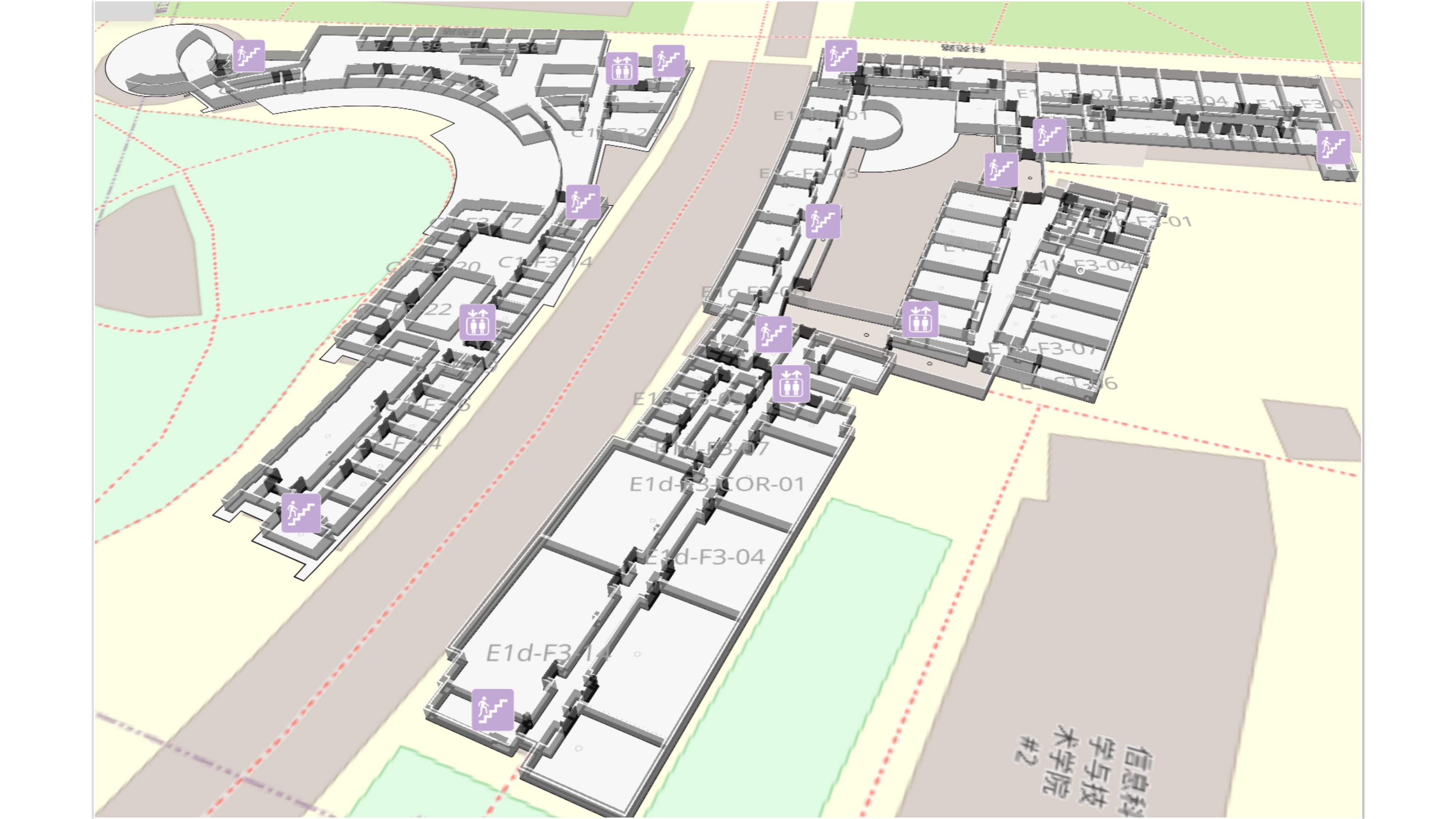}
    \caption{\textbf{Rendered example of an osmAG map in an OSM-compatible indoor renderer.} Rooms, corridors, and larger structural regions are visualized as vector polygons over a georeferenced base layer, while elevator and stair icons expose the vertical-transition semantics directly in the map view. The rendering illustrates that osmAG is not merely a storage format, but a semantic-topological representation that can be inspected, validated, and edited within the OSM ecosystem.}
    \label{fig:osmag_rendered}
\end{figure*}

A rendered osmAG floor is shown in Fig.~\ref{fig:osmag_rendered}.

\subsection{Canonical OSM/XML Schema}
\label{subsec:osmag_schema}

Table~\ref{tab:osmag_core_schema} condenses the minimum canonical field set required to exchange a valid core osmAG map. The semantic \texttt{name} of each area and passage is the canonical reference target for relational tags. Consequently, names must be globally unique within a single osmAG file. This choice keeps the exchange format stable under manual editing in tools such as JOSM, where raw OSM way identifiers may be reassigned during save operations.

\begin{table}[t]
\caption{Minimum canonical field set of the core osmAG standard.}
\label{tab:osmag_core_schema}
\centering
\scriptsize
\setlength{\tabcolsep}{3pt}
\renewcommand{\arraystretch}{1.0}
\begin{tabularx}{0.98\columnwidth}{@{}>{\raggedright\arraybackslash}p{0.28\columnwidth}>{\raggedright\arraybackslash}p{0.13\columnwidth}>{\centering\arraybackslash}p{0.10\columnwidth}X@{}}
\toprule
\textbf{Field} & \textbf{Applies to} & \textbf{Req.} & \textbf{Meaning} \\
\midrule
\texttt{id, lat, lon} & node & Yes & OSM-native geometric node definition in WGS84. \\
\texttt{name=root} & node & Yes (once) & Unique root anchor for the local Cartesian frame. \\
\texttt{name} & area, passage & Yes & Globally unique semantic identifier; canonical target of references. \\
\texttt{osmAG:type} & area, passage & Yes & Declares \texttt{area} or \texttt{passage}. \\
\texttt{osmAG:areaType} & area & Yes & Canonical area class: \texttt{room}, \texttt{corridor}, \texttt{structure}, \texttt{elevator}, or \texttt{stairs}. \\
\texttt{osmAG:parent} & area & Cond. & Parent area name; required for non-top-level areas. \\
\texttt{osmAG:from}, \texttt{osmAG:to} & passage & Yes & Incident area names; unordered in the core standard. \\
\texttt{level} & area, passage & Cond. & Floor index; required for indoor multi-floor structure. \\
\texttt{height} & area, passage & No & Optional elevation value following OSM convention. \\
\texttt{indoor} & area & No & OSM interoperability tag used by authoring and visualization tools. \\
\botrule
\end{tabularx}
\end{table}

\subsection{Hierarchy and Map Validity Constraints}
\label{subsec:osmag_validity}

An osmAG file is valid only if the following invariants hold:
\begin{enumerate}
    \item \textbf{Tree invariant.} Every non-top-level area has at most one parent area, and hierarchical references form a forest of containment trees rather than arbitrary cycles.
    \item \textbf{Containment invariant.} If an area $A_c$ references a parent area $A_p$, then the geometry of $A_c$ must lie within the geometry of $A_p$.
    \item \textbf{Geometric consistency invariant.} Two areas sharing the same immediate parent and the same \texttt{level} must have disjoint interiors. Overlap is permitted only between areas on different floors (vertical stacking) or between an ancestor and its descendants (containment).
    \item \textbf{Passage adjacency invariant.} Every passage must connect exactly two existing named areas and must encode a traversable boundary or transfer relation between them.
\end{enumerate}

These invariants are not merely descriptive. They provide the validity assumptions used later by hierarchical planning, which relies on well-formed containment to lift search through the map hierarchy, and by structure-based localization, which relies on persistent area geometry to define stable structural priors.

\subsection{Vertical Topology Standard}
\label{subsec:osmag_vertical}

osmAG treats elevators and stairs as standard areas with specialized semantics rather than as planner-side exceptions. A floor-specific elevator or stair footprint is encoded as an area with \texttt{osmAG:areaType=elevator} or \texttt{osmAG:areaType=stairs} and is typically parented to the floor-level structure area that contains it.

Two passage roles are distinguished:
\begin{itemize}
    \item \textbf{Floor-access passages} connect a vertical area to neighboring areas on the same floor, for example from an elevator lobby to the elevator shaft footprint.
    \item \textbf{Inter-floor passages} explicitly connect the floor-specific vertical areas that represent the same physical shaft or stairwell across adjacent floors.
\end{itemize}

Because each floor-specific vertical area carries a unique \texttt{name}, vertical continuity is encoded explicitly by inter-floor passages rather than by reusing a single area identifier across floors. This makes vertical transfer a first-class part of the map topology and allows multi-floor navigation to be formulated as ordinary graph search over standardized map elements.

\subsection{Sanctioned Semantic Extension Profiles}
\label{subsec:osmag_extensions}

The core standard intentionally keeps the normative geometry and hierarchy compact. Richer task semantics are layered on top through sanctioned extension profiles that preserve the same OSM/XML backbone.

\subsubsection{Navigation Semantics Profile}

The navigation profile standardizes optional task metadata under the \texttt{osmAG:*} namespace. Typical keys include \texttt{osmAG:area\_usage} for functional area roles, \texttt{osmAG:room\_number} for human-facing room identifiers, \texttt{osmAG:occupied\_by} for stable ownership or tenant metadata, and \texttt{osmAG:passage\_type} for passage semantics such as automatic or handle-operated doors. These tags refine routing behavior and human readability without changing the underlying topology. In contrast, \texttt{osmAG:degree} is treated only as optional simulation metadata for articulated doors and is not part of the core map standard.

\subsubsection{Embodied-AI Semantic Profile}

The Embodied-AI profile retains the dedicated \texttt{semantic\_osmAG:*} namespace for semantic enrichment beyond classical navigation. In this profile, object nodes and viewpoint nodes are stored as ordinary OSM nodes anchored in the same reference frame as the base map, while areas may carry higher-level language descriptions. Canonical sanctioned keys are \texttt{semantic\_osmAG:object\_name} for named object instances with explicit positions, \texttt{semantic\_osmAG:observed\_objects} for viewpoint-based open-vocabulary observations, and \texttt{semantic\_osmAG:area\_description} for room-level natural-language summaries. This profile is designed to support downstream reasoning systems such as zero-shot object navigation and semantic retrieval without bloating the core map specification.

Because osmAG preserves the standard OSM node/way data model, canonical files can be authored and inspected with mainstream OSM-compatible tooling such as JOSM and visualized in OpenIndoor. Legacy spellings (e.g., \texttt{osmAG:areatype}) and older ID-based file variants are supported only through conversion utilities and are not part of the canonical exchange format. A fuller tag-level reference for all core and extension fields is provided in Appendix~\ref{sec:appendix_osmag_reference}; quantitative storage comparisons are reported in Section~\ref{sec:experiments}.

%% file: sections/4_sys_architecture.tex
\section{System Architecture Overview}
\label{sec:sys_architecture}

\begin{figure*}[t]
    \centering
    \includegraphics[width=\linewidth]{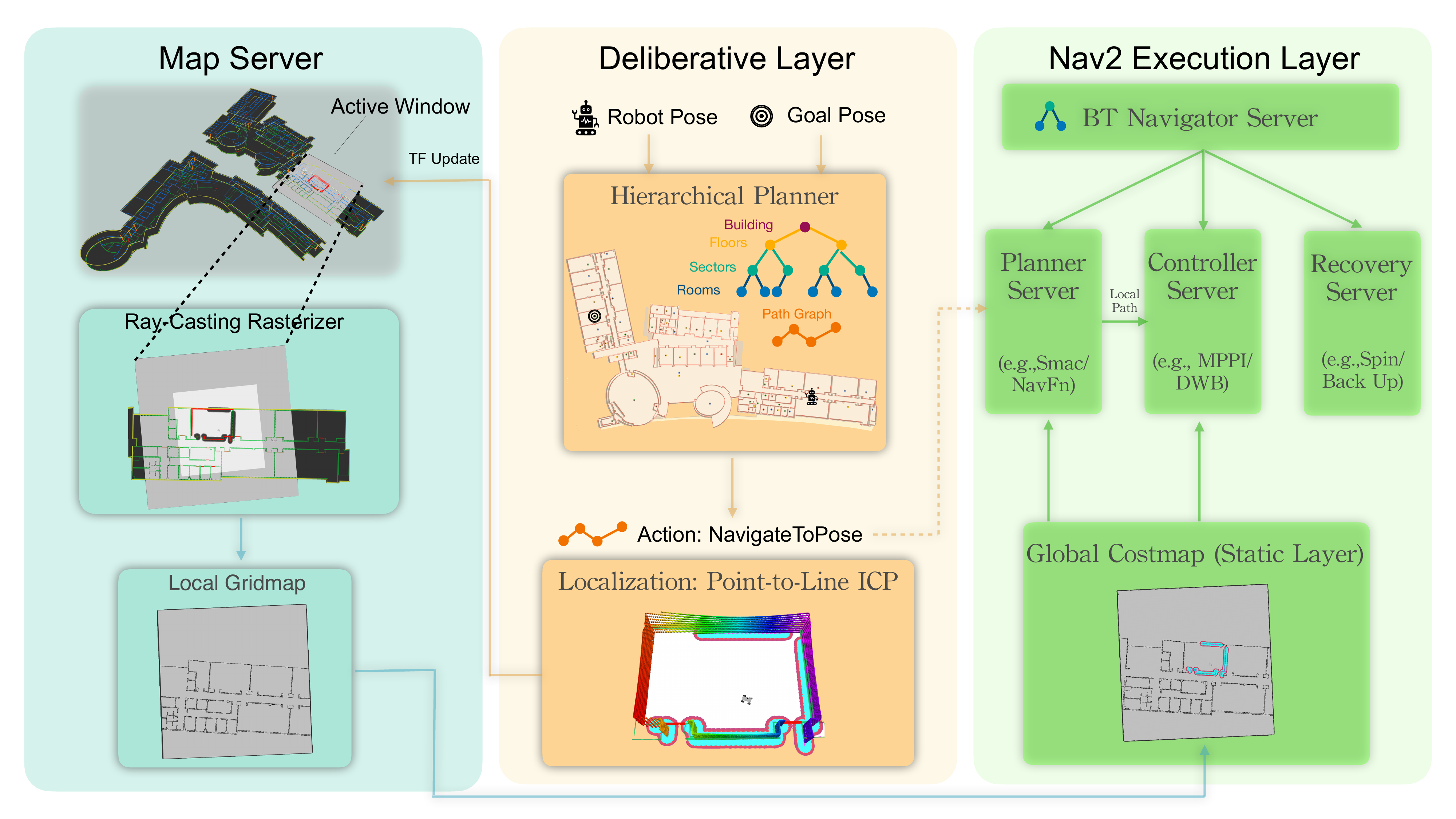}
    \caption{\textbf{System architecture of osmAG-Nav.} The stack follows a decoupled ``system of systems'' design. The environment layer maintains the global semantic-topological map and serves a bounded Rolling Window raster through \texttt{/local\_AGgridmap}. The deliberative layer combines structure-based localization with hierarchical passage-centric planning in the global semantic frame. The execution layer remains standard Nav2, consuming the injected metric map and intermediate goals through ROS2-native interfaces.}
    \label{fig:system_architecture}
\end{figure*}

osmAG-Nav is implemented in ROS2 as a set of cooperating subsystems rather than as a monolithic navigation node. This design is deliberate. We preserve the robustness and modularity of the standard Nav2 execution pipeline, but replace the global world model and the deliberative reasoning layer with an osmAG-based semantic-topological stack. As a result, long-horizon reasoning is performed on a sparse hierarchical vector map, while short-horizon motion is still executed in a bounded metric space.

\subsection{Overview and Design Philosophy}
\label{subsec:overview_philosophy}

The architecture is organized around three principles.

\begin{itemize}
    \item \textbf{Persistent global structure, ephemeral local metric state.} The full environment is maintained as a lightweight semantic-topological vector map, while only a bounded local metric patch is exposed to downstream planners. Unlike a monolithic global costmap, the map consumed by Nav2 remains independent of the total building scale.

    \item \textbf{Topological deliberation, metric execution.} Global routing is handled by the osmAG planner at the passage level. Local geometry, collision avoidance, and kinodynamic feasibility are delegated to standard Nav2 components. In contrast to a monolithic planner-controller stack, this preserves long-horizon topological control without reimplementing mature local execution modules.

    \item \textbf{ROS2-native interface compatibility.} The system communicates through standard ROS2 topics, actions, and TF relations. The Rolling Window map is published as \texttt{nav\_msgs/OccupancyGrid}, segment goals are dispatched through the \texttt{NavigateToPose} action, and localization is exposed through Nav2-compatible pose and TF interfaces.
\end{itemize}

\subsection{Layered Organization}
\label{subsec:ros2_implementation}

The deployed system is organized into three layers rather than around a single planner-centric process.

\subsubsection{Environment Layer}

The environment layer maintains the persistent semantic-topological world model and exposes the bounded local metric view needed by the execution stack. In our implementation, this layer is realized by the osmAG parser/publisher together with the Rolling Window map server. These components provide structural map data for localization, topological state for planning, and the local occupancy grid injected into Nav2.

\subsubsection{Deliberative Layer}

The deliberative layer combines global topological reasoning with structure-based state estimation. The planner operates on the passage-centric osmAG graph and manages mission progress through segmented goal dispatch. In parallel, the localization subsystem estimates pose against permanent architectural structure and exposes the resulting state through Nav2-compatible interfaces. This layer therefore owns the long-horizon route, the topological interpretation of the environment, and the global frame used for reasoning.

\subsubsection{Execution Layer}

The execution layer is deliberately conventional. Standard Nav2 components consume the injected local grid map, receive segment goals through \texttt{NavigateToPose}, and handle short-horizon path generation and control. Unlike systems that embed local control logic into the global planner itself, osmAG-Nav preserves the mature local planning and control pipeline of Nav2 and changes only the representation and reasoning layers above it.

\subsection{Frames and Interface Boundaries}
\label{subsec:data_flow}

The stack is organized around three distinct frame roles.

\begin{itemize}
    \item \textbf{\texttt{AGmap}} is the semantic-topological global frame. The planner computes and publishes its global route in this frame, and the localization core also estimates pose relative to it.
    \item \textbf{\texttt{map}} is the metric navigation frame consumed by Nav2. The Rolling Window map is published in this frame, and \texttt{map $\rightarrow$ AGmap} links the local metric view to the global semantic map.
    \item \textbf{\texttt{odom}} provides the short-term motion frame used by the execution stack, with \texttt{map $\rightarrow$ odom} supplied by the localization adapter and \texttt{odom $\rightarrow$ base\_link} supplied by odometry or LiDAR-inertial estimation.
\end{itemize}

Three interfaces are architecturally critical:

\begin{itemize}
    \item \textbf{\texttt{/local\_AGgridmap}} links the environment layer to Nav2 by exposing a bounded metric map.
    \item \textbf{\texttt{NavigateToPose}} links the deliberative layer to the execution layer by turning a global topological route into intermediate metric goals.
    \item \textbf{TF and pose publication} link localization to both planning and execution by reconciling \texttt{AGmap}, \texttt{map}, and \texttt{odom}.
\end{itemize}

This separation is central to the design. Nav2 is not required to reason directly over the full semantic-topological map, and the planner is not forced to operate in a purely local metric frame. Instead, ROS2 topics, actions, and TF provide the coupling between layers while preserving their distinct responsibilities.

\subsection{End-to-End Data Flow}

At runtime, the pipeline proceeds as follows.

\begin{enumerate}
    \item The osmAG parser loads the vector map, publishes structural map data for localization, and initializes the semantic-topological representation used by planning.

    \item The localization subsystem estimates robot pose against permanent structural priors and exposes the result through Nav2-compatible pose and TF interfaces.

    \item The Rolling Window server uses the current pose and floor context to generate a bounded local occupancy grid and publishes it to the Nav2 static layer.

    \item The hierarchical planner computes a global passage-level route in \texttt{AGmap}, selects the active segment, and dispatches an intermediate goal through \texttt{NavigateToPose}.

    \item Nav2 plans and controls within the local metric window, while the planner monitors segment completion and advances the mission state to the next passage transition.
\end{enumerate}

The architectural contribution lies not merely in the coexistence of these components, but in the way they are coupled through stable interfaces. Unlike a purely occupancy-grid stack, osmAG-Nav reasons globally over semantic topology. Unlike a purely topological navigator, it still inherits the mature local planning and control capabilities of Nav2.

%% file: sections/5_navigation.tex
\section{Hierarchical Navigation Methodology}
\label{sec:nav}

This section describes how \textit{osmAG-Nav} converts a start-goal query on a persistent semantic-topological map into executable motion. Building upon the interface boundaries established in Section~\ref{sec:sys_architecture}, the method combines four tightly coupled mechanisms: on-demand Rolling Window rasterization as a topology-to-metric bridge, passage-centric graph construction with physically-aware raster edge costs, hierarchical search following \textit{attach} $\rightarrow$ \textit{lift} $\rightarrow$ \textit{common-parent compact-graph A*} $\rightarrow$ \textit{expand}, and segmented execution through Nav2.

\subsection{Rolling Window as a Topology-to-Metric Bridge}
\label{subsec:rolling_window}

Nav2 consumes occupancy grids, whereas osmAG stores a sparse vector hierarchy. We bridge this representational gap by maintaining only a fixed-size local window in the robot vicinity and publishing it as \texttt{/local\_AGgridmap}. In the deployed implementation, the window size is $50\,\mathrm{m} \times 50\,\mathrm{m}$ and the update loop runs at $10\,\mathrm{Hz}$. Because the window size is fixed, the local costmap memory footprint remains bounded independently of the total mapped area, yielding $O(1)$ local-map memory with respect to environment scale.

The Rolling Window server is an online bridge from vector topology to metric execution rather than a pre-rendered floor map. At each update, the current implementation first scans the stored area set to retain floor-consistent polygons whose boundary nodes intersect the active window. It then scans only the grid bounding box of each retained polygon and applies a ray-casting point-in-polygon test to mark free cells. Non-passage boundaries are then drawn as occupied, whereas passage segments are explicitly reopened as free space. Accordingly, the per-update rasterization cost scales with the number of candidate areas intersecting the window and, for each retained polygon, with the size of its scanned bounding box and its vertex count. This online rasterization path should be distinguished from the separate OpenCV-based rasterization pipeline used offline for leaf-area distance precomputation.

\begin{figure}[htbp]
    \centering
    \includegraphics[width=0.95\columnwidth]{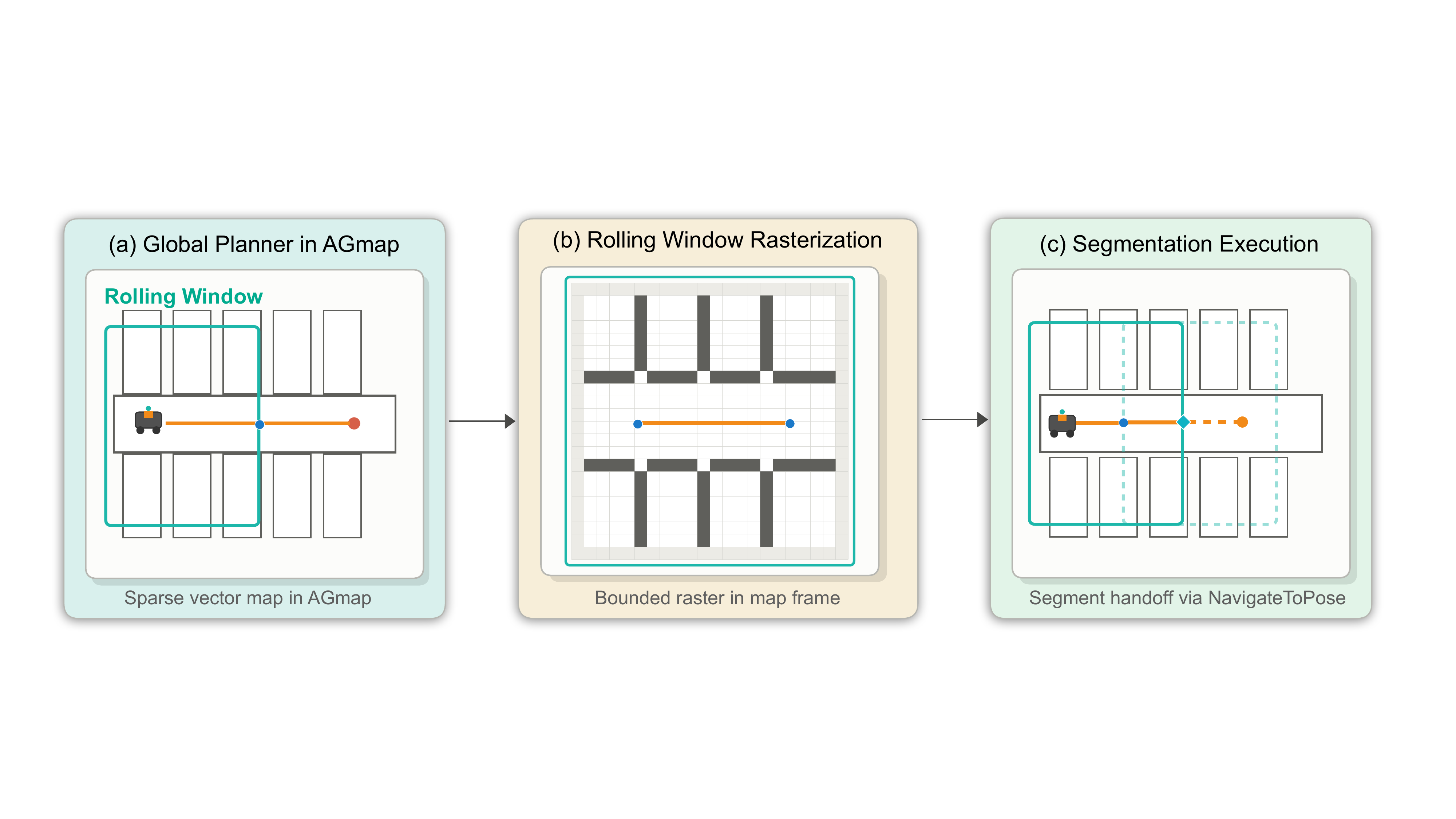}
    \caption{\textbf{Topology-to-Metric Bridging with Rolling Window and Segmented Execution.} (a) The planner reasons over the full semantic-topological map in \texttt{AGmap}, while only a bounded Rolling Window is rasterized and published as \texttt{/local\_AGgridmap} in the \texttt{map} frame for Nav2. (b) Within the local window, area interiors are filled as free space, non-passage boundaries are marked occupied, and passage segments are explicitly reopened to preserve traversability. (c) The planner dispatches intermediate passage goals through \texttt{NavigateToPose}; when the desired passage lies beyond the current local map, a proxy goal is projected to the local boundary and refreshed as the window advances.}
    \label{fig:segmented_execution}
\end{figure}

The resulting occupancy grid is published in the \texttt{map} frame with \texttt{transient\_local} QoS and consumed by Nav2 through its \texttt{StaticLayer}. The planner itself does not search on this raster. Instead, global reasoning remains in \texttt{AGmap}, while the Rolling Window provides only the bounded metric support required for execution and, later, for local-map-aware goal projection. The bridge is therefore deliberately asymmetric: the global map stays sparse and persistent, whereas the metric map is local and ephemeral. Figure~\ref{fig:segmented_execution} summarizes this topology-to-metric coupling together with the segmented execution logic built on top of it.

\subsection{Physically-Aware Graph Construction}
\label{subsec:graph_construction}

Global routing is performed over passages rather than room centroids or grid cells. The core search structure is a sparse passage graph $\mathcal{G} = (\mathcal{V}, \mathcal{E})$ whose vertices $\mathcal{V}$ denote transition elements such as doors, corridor openings, stairs, and elevator interfaces. This passage-centric representation keeps the search state aligned with the bottlenecks that actually constrain motion through buildings.

Each leaf area is treated as the basic unit of direct traversability from which the base passage graph is derived offline. For every leaf area $A_k$, the planner instantiates its resident passages as graph vertices located at their geometric centers and connects only those passage pairs that represent distinct transitions through the same area, while redundant pairs induced by the same adjacent-area interface are discarded. For planar leaf areas, the edge cost and the associated intra-area trace are computed on a local occupancy raster via grid A*, whereas stairs and elevator transitions bypass planar raster search and receive the fixed parameter $c_{\mathrm{vert}}$; Euclidean connection is used only as a fallback when raster search fails. The union of these leaf-level edges forms the persistent global passage graph, from which area compact graphs and parent lift graphs are cached as hierarchical summaries. At query time, the start and goal poses are localized to their containing leaf areas, inserted temporarily as virtual passages, and attached to resident passages through the same leaf-area rasterization pipeline. This temporary augmentation grounds arbitrary queries in geometry without altering the persistent passage graph across planning episodes.

Algorithm~\ref{alg:base_passage_graph} summarizes the offline construction of the persistent passage graph that underlies the subsequent hierarchical search.

\begin{algorithm}[t]
\caption{Offline Construction of the Base Passage-Centric Graph}
\label{alg:base_passage_graph}
\begin{algorithmic}[1]
\Require osmAG hierarchy $\mathcal{H}$ with leaf areas $\mathcal{A}_{\mathrm{leaf}}$
\Ensure Base passage graph $\mathcal{G}_b=(\mathcal{V},\mathcal{E})$ and intra-area trace set $\mathcal{T}$
\State $\mathcal{V} \gets \emptyset,\ \mathcal{E} \gets \emptyset,\ \mathcal{T} \gets \emptyset$
\ForAll{$A_k \in \mathcal{A}_{\mathrm{leaf}}$}
    \State $\mathcal{P}_k \gets \Call{ResidentPassages}{A_k}$
    \ForAll{$p \in \mathcal{P}_k$}
        \State $\mathcal{V} \gets \mathcal{V} \cup \{p\}$
    \EndFor
    \If{$A_k$ is planar}
        \State $M_k \gets \Call{RasterizeLeafArea}{A_k}$
    \EndIf
\ForAll{unordered passage pairs $(p_i,p_j) \subset \mathcal{P}_k$}
        \If{$\Call{SameAdjacentAreaInterface}{p_i,p_j}$}
            \State \textbf{continue}
        \EndIf
        \If{$\Call{VerticalTransition}{A_k,p_i,p_j}$}
            \State $w_{ij} \gets c_{\mathrm{vert}},\ \tau_{ij} \gets [p_i,p_j]$
        \Else
            \State $(w_{ij}, \tau_{ij}) \gets \Call{GridAStar}{M_k,p_i,p_j}$
            \If{$\tau_{ij}$ invalid}
                \State $w_{ij} \gets \|p_i-p_j\|_2,\ \tau_{ij} \gets [p_i,p_j]$
            \EndIf
        \EndIf
        \State $\mathcal{E} \gets \mathcal{E} \cup \{(p_i,p_j,w_{ij})\}$
        \State $\mathcal{T}[p_i,p_j] \gets \tau_{ij}$
    \EndFor
\EndFor
\State $\mathcal{G}_b \gets (\mathcal{V},\mathcal{E})$
\State $\Call{BuildAreaCompactGraphs}{\mathcal{G}_b}$
\State $\Call{BuildParentLiftGraphs}{\mathcal{G}_b}$
\State \Return $\mathcal{G}_b=(\mathcal{V},\mathcal{E}),\ \mathcal{T}$
\end{algorithmic}
\end{algorithm}

Pure Euclidean edge weights are unreliable indoors because walls, columns, and narrow openings can force substantial detours within an area. We therefore assign \textbf{physically-aware raster edge costs} by precomputing passage-to-passage traversal costs on a leaf-area occupancy raster:
\begin{equation}
    w_{ij} =
    \begin{cases}
        \mathrm{RasterAStarCost}_{A_k}(p_i, p_j), & \text{non-vertical edge in leaf area } A_k, \\
        c_{\mathrm{vert}}, & \text{vertical transition edge}, \\
        \|p_i - p_j\|_2, & \text{fallback if local raster search fails}.
    \end{cases}
\end{equation}
In the deployed implementation, $\mathrm{RasterAStarCost}_{A_k}$ is obtained by rasterizing each leaf area at $0.1\,\mathrm{m}$ resolution and running grid A* between passage pairs. The resulting value should be interpreted as a geometry-aware raster traversal cost that preserves obstacle-induced detours, rather than as a full kinodynamic-feasibility certificate or an exact continuous-space path length in meters. Vertical transitions bypass planar raster search and instead receive the fixed parameter \texttt{vertical\_transition\_cost}; if a local search fails, the implementation falls back to Euclidean connection. This design preserves the sparsity of the topological graph while making its edge costs substantially more faithful to traversable indoor geometry than straight-line weights.

\subsection{Attach-Lift-Common-Parent Hierarchical Planning}
\label{subsec:hierarchical_search}

The online planner is not a generic ``reduced graph A*'' over an ad hoc subgraph. After inserting temporary start and goal passages, it first locates their containing leaf areas, computes the lowest common ancestor (common parent), and then executes the query-specific sequence
\[
\textit{attach} \rightarrow \textit{lift} \rightarrow \textit{common-parent compact-graph A*} \rightarrow \textit{expand}.
\]
During initialization, compact graphs are built for leaf areas and parent lift graphs are cached to summarize inter-child connectivity higher in the hierarchy. These structures are then reused across queries.

\begin{figure}[htbp]
    \centering
    \includegraphics[width=0.95\columnwidth]{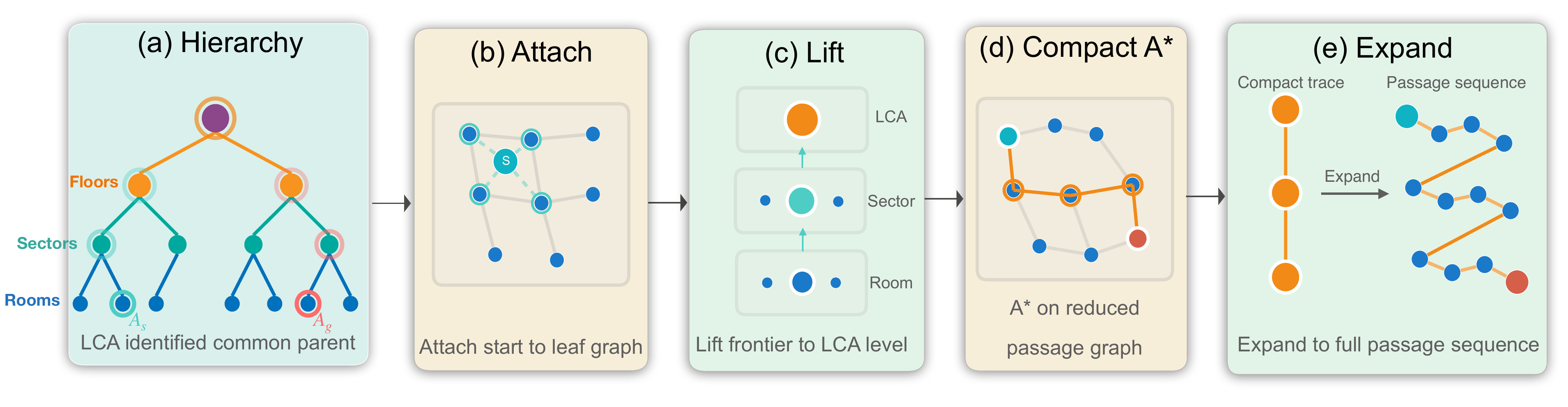}
    \caption{\textbf{Attach-Lift-Common-Parent Hierarchical Planning.} (a) The osmAG hierarchy (Building $\rightarrow$ Floors $\rightarrow$ Sectors $\rightarrow$ Rooms) with start area $A_s$ and goal area $A_g$ highlighted, showing their ancestry paths to the common parent $A_c$. (b) The planner injects virtual passage $s$ and attaches it to the leaf compact graph via multi-source Dijkstra. (c) The frontier is lifted upward through parent lift graphs until reaching the common parent. (d) A* runs on the assembled common-parent compact graph to find the compact trace. (e) The compact trace is expanded back to the full passage sequence.}
    \label{fig:hierarchical_engine}
\end{figure}

\begin{algorithm}[t]
\caption{Attach-Lift-Common-Parent Hierarchical Planning}
\label{alg:hierarchical_planning}
\begin{algorithmic}[1]
\Require Start pose $\mathbf{x}_s$, goal pose $\mathbf{x}_g$, osmAG hierarchy $\mathcal{H}$
\Ensure Passage sequence $\mathcal{P}$
\State $(s, A_s) \gets \Call{InjectVirtualPassage}{\mathbf{x}_s}$
\State $(g, A_g) \gets \Call{InjectVirtualPassage}{\mathbf{x}_g}$
\State $A_c \gets \Call{LowestCommonAncestor}{A_s, A_g}$
\State $\phi \gets \Call{SameFloorConstraintIfApplicable}{s, g}$
\State $\mathcal{F}_s \gets \Call{Attach}{s, A_s, \phi}$ \Comment{multi-source Dijkstra on leaf compact graph}
\State $\mathcal{F}_g \gets \Call{Attach}{g, A_g, \phi}$
\If{$\mathcal{F}_s$ or $\mathcal{F}_g$ invalid}
    \State \Return $\Call{GlobalPassageAStar}{s, g}$
\EndIf
\While{$A_s \neq A_c$ \textbf{ and } $\Call{Parent}{A_s} \neq A_c$}
    \State $(\mathcal{F}_s, A_s) \gets \Call{Lift}{\mathcal{F}_s, A_s, \phi}$ \Comment{multi-source Dijkstra on parent lift graph}
    \If{$\mathcal{F}_s$ invalid}
        \State \Return $\Call{GlobalPassageAStar}{s, g}$
    \EndIf
\EndWhile
\While{$A_g \neq A_c$ \textbf{ and } $\Call{Parent}{A_g} \neq A_c$}
    \State $(\mathcal{F}_g, A_g) \gets \Call{Lift}{\mathcal{F}_g, A_g, \phi}$
    \If{$\mathcal{F}_g$ invalid}
        \State \Return $\Call{GlobalPassageAStar}{s, g}$
    \EndIf
\EndWhile
\State $\mathcal{G}_c \gets \Call{AssembleCommonParentCompactGraph}{A_c, \mathcal{F}_s, \mathcal{F}_g, \phi}$
\If{$\mathcal{G}_c$ invalid}
    \State \Return $\Call{GlobalPassageAStar}{s, g}$
\EndIf
\State $\mathcal{P}_{compact} \gets \Call{AStar}{\mathcal{G}_c, s, g}$
\State $\mathcal{P} \gets \Call{ExpandCompactTrace}{\mathcal{P}_{compact}, \phi}$
\If{$\mathcal{P}$ invalid}
    \State $\mathcal{P} \gets \Call{GlobalPassageAStar}{s, g}$
\EndIf
\State \Return $\mathcal{P}$
\end{algorithmic}
\end{algorithm}

Two implementation details are critical. First, the attach and lift stages are both realized as multi-source Dijkstra propagation on cached compact graphs rather than as repeated full-graph search. Second, the two propagated frontiers are not simply concatenated once they approach the common parent. Instead, the planner assembles a \emph{common-parent compact planning graph} by combining the cached lift graph at the common parent with the traced attach and lift edges from both sides, runs A* once on that compact graph, and only then expands the compact result back to the underlying passage sequence. If any key stage fails, the implementation falls back to global passage-level A*.

This organization makes the search volume depend on the relevant ancestry around the query rather than on the full passage graph. The hierarchy is therefore used as an online planning substrate rather than merely as a descriptive map structure. In addition, when both endpoint passages carry the same floor tag, the planner switches to a same-floor-only mode during attach, lift, and common-parent graph assembly so that unnecessary vertical detours are excluded before compact A* is even attempted. Cross-floor routes remain available when the endpoint floor tags differ.

\subsection{Segmented Execution with Local-Map Projection}
\label{subsec:execution}

The hierarchical planner outputs both an expanded passage sequence and a dense geometric path in \texttt{AGmap}. Execution is deliberately segmented: rather than handing Nav2 a single terminal destination, the planner retains high-level topological control, publishes the full \texttt{osmag\_path}, and dispatches intermediate goals through the standard \texttt{NavigateToPose} action interface.

Each segment goal is placed at the center of its target passage and augmented with an orientation cue so that local execution does not treat passage crossings as isolated points. In the deployed implementation, this heading is derived from the vector from the selected passage toward the final goal, which biases the approach toward downstream route continuation without requiring a fully pose-annotated topological map.

The planner monitors the distance between the robot and the active segment goal. Once the robot enters a configurable handoff zone defined by \texttt{goal\_reach\_threshold}, the next \texttt{NavigateToPose} goal is issued without waiting for a full stop at the current passage. This continuous handoff mechanism preserves motion continuity while keeping the high-level route under planner control.

\subsubsection{Boundary-Aware Goal Projection}
Because Nav2 only sees the latest Rolling Window map, the next desired passage may lie outside the currently available metric bounds. In that case, the planner does not relinquish control to Nav2. Instead, it checks candidate goals across the \texttt{AGmap}--\texttt{map} boundary, first searching along the already generated \texttt{osmag\_path} for the farthest pose that still lies inside the latest \texttt{/local\_AGgridmap} boundary. If no such pose can be recovered, it falls back to a bounded projection inside the local map with a safety margin. The resulting proxy goal is transformed back to \texttt{AGmap} and dispatched through \texttt{NavigateToPose}, while Nav2 continues to execute in its usual \texttt{map}/\texttt{odom} frame chain. As shown in Fig.~\ref{fig:segmented_execution}, this projection logic is part of the same bridge that couples the persistent global route to bounded local metric execution.

%% file: sections/6_localization.tex
\section{Robust Localization Framework}
\label{sec:loc}

\begin{figure}[htbp]
    \centering
    \includegraphics[width=0.95\columnwidth]{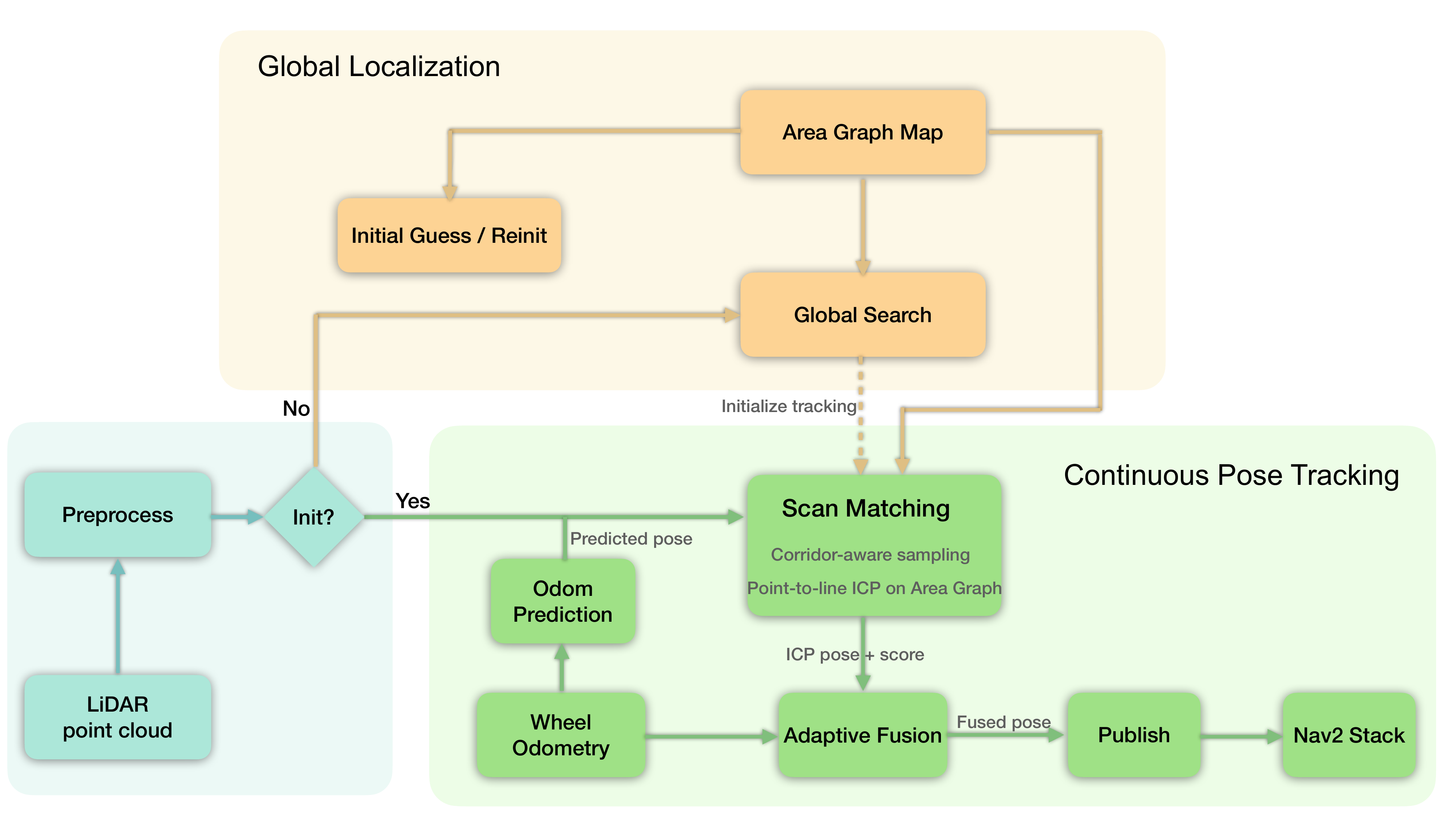}
    \caption{\textbf{Localization Robustness Pipeline.} Preprocessed LiDAR observations first pass through an initialization check. If the system is uninitialized or requires recovery, the global-localization branch uses the Area Graph map together with an initial guess or reinitialization prior to perform global search and to bootstrap tracking. Once initialized, the system enters continuous pose tracking, where corridor-aware scan matching and odometry prediction run in parallel, the resulting ICP pose and matching score are fused adaptively with wheel odometry, and the fused pose is published to the Nav2 stack.}
    \label{fig:localization_framework}
\end{figure}

Building upon the AGLoc framework introduced in our prior work~\citep{xie2023robust}, the localization subsystem in \textit{osmAG-Nav} is redesigned as a structure-first estimator for long-term deployment. Its role in this paper is not to compete with the planner for methodological focus, but to provide the state-estimation backbone that keeps the semantic-topological stack anchored to the physical environment. The core estimator matches LiDAR observations against permanent architectural structure encoded by the Area Graph, while the Nav2-facing adapter exposes the resulting state through standard pose and TF interfaces. As shown in Fig.~\ref{fig:localization_framework}, the pipeline combines optional global relocalization, structure-based scan-to-map tracking, corridor-aware constraint selection, and adaptive fusion with odometric prediction. In the experimental evaluation (Section~\ref{sec:experiments}), this enhanced configuration is referred to as \textbf{AGLoc++} to distinguish it from the original AGLoc baseline.

\subsection{Global Initialization and Relocalization}
\label{subsec:global_reloc}

The implementation supports two operating paths. In the common case, once a consistent pose has been established, the system remains in continuous tracking mode. When the robot is uninitialized or recovery is requested, the rescue branch performs global localization before handing control back to the tracker. In the runtime profiles used for end-to-end navigation, tracking is the default path, while global relocalization remains available as an explicit recovery capability rather than a per-frame burden.

Instead of relying on a single exhaustive search over the full pose space, the initialization module uses a staged procedure. Candidate particles are first sparsified, then evaluated by an area-wise coarse angular search, after which the two strongest modes are refined locally and verified by ICP. This design keeps relocalization tractable while reducing the risk of committing prematurely to one ambiguous corridor hypothesis. For a candidate pose, the base geometric consistency score is computed as
\begin{equation}
    S_{\mathrm{base}} = \frac{1}{S_{\mathrm{inside}} + S_{\mathrm{outside}}}, \qquad
    S_{\mathrm{final}} = S_{\mathrm{base}} \cdot \eta_{\mathrm{edge}},
\end{equation}
where $S_{\mathrm{inside}}$ and $S_{\mathrm{outside}}$ denote the accumulated residuals of points that are geometrically consistent with the mapped area interior and exterior, respectively, and $\eta_{\mathrm{edge}}$ penalizes weakly supported solutions, such as hypotheses with poor inside/outside balance, excessive exterior support, or boundary-like alignment. The resulting relocalization branch is therefore guided by structural consistency rather than by raw scan overlap alone.

\subsection{Map-Based Clutter Removal and Point-to-Line Tracking}
\label{subsec:clutter_tracking}

After initialization, every incoming scan is filtered against the structural map before optimization. For a LiDAR return $\mathbf{p}_i$, the system evaluates both a ray-consistency term
\begin{equation}
    e_i = d_i^{\mathrm{scan}} - d_i^{\mathrm{map}},
\end{equation}
and the perpendicular distance $d_{\perp}(\mathbf{p}_i, \mathbf{L}_i)$ to the matched map segment $\mathbf{L}_i$. A point is retained only if it is geometrically compatible with the permanent structure:
\begin{equation}
    \mathcal{P}_{\mathrm{clean}} =
    \left\{
    \mathbf{p}_i \in \mathcal{P}_{\mathrm{raw}}
    \,\middle|\,
    \bigl(e_i \leq 0 \land d_{\perp}(\mathbf{p}_i, \mathbf{L}_i) < \tau_{\mathrm{in}}\bigr)
    \,\vee\,
    \bigl(e_i > 0 \land d_{\perp}(\mathbf{p}_i, \mathbf{L}_i) < \tau_{\mathrm{out}}\bigr)
    \right\},
\end{equation}
where $\tau_{\mathrm{in}}$ and $\tau_{\mathrm{out}}$ are the inside/outside gating thresholds used by the tracker. This map-based clutter removal rejects pedestrians, moved furniture, and wall-penetrating reflections before they can corrupt the optimizer.

Tracking is then formulated as a point-to-line ICP problem,
\begin{equation}
    \mathbf{T}^{*} = \arg\min_{\mathbf{T}} \sum_i w_i \, d_{\perp}(\mathbf{T}\mathbf{p}_i,\mathbf{L}_i)^2,
\end{equation}
so that the scan is corrected directly against the persistent structural map rather than against short-term scan history. To improve robustness, the implementation computes an asymmetric robust weight for each accepted point:
\begin{equation}
    w_i^{\mathrm{rob}} =
    \begin{cases}
        \dfrac{\tau_{\mathrm{out}}}{9r_i + \tau_{\mathrm{out}}}, & \text{outside-supported point}, \\
        \dfrac{\tau_{\mathrm{in}}}{1.5r_i + \tau_{\mathrm{in}}}, & \text{inside-supported point}, \\
        0, & \text{otherwise},
    \end{cases}
\end{equation}
where $r_i$ is the point-to-line residual. Outside-supported points are downweighted more aggressively because they are more likely to arise from glass, clutter, or partial wall penetration. The implementation always computes these robust weights; whether they are injected directly into the ICP accumulation is kept configuration-dependent, which cleanly separates the robustness machinery from the chosen runtime solve policy.

\subsection{Corridor-Aware Tracking Stability}
\label{subsec:corridor_stability}

Long corridors are a classical degeneracy case because many observations support one dominant direction while providing weak constraint along the corridor axis. The estimator therefore builds an orientation histogram over matched structural lines to quantify \emph{corridorness}. When one direction becomes over-represented, observations aligned with that direction are adaptively downsampled, whereas measurements that remain informative in the map-normal direction are explicitly preserved. In other words, the tracker applies direction-aware geometric constraints that selectively retain near-orthogonal features instead of allowing the dominant corridor direction to overwhelm the update. This preference can be written as
\begin{equation}
    \tilde{w}_i = w_i^{\mathrm{rob}} \cdot \mathrm{clamp}\!\left(\left|\mathbf{n}_i^{\top}\mathbf{v}_i\right|, 0.3, 1.0\right),
\end{equation}
where $\mathbf{n}_i$ is the normal of the matched map segment and $\mathbf{v}_i$ is the observation direction from the robot to the point. The additional factor strengthens cross-axis constraints and suppresses the axial drift and aliasing modes that are typical in elongated, feature-poor hallways.

Robustness is reinforced further at the update level. Before each ICP increment is applied, excessive translation or rotation steps are clamped to configurable maxima rather than being accepted blindly. This dynamic step-limiting mechanism serves as explicit optimization divergence protection, and a subsequent frame-level jump guard rejects implausible pose excursions that remain inconsistent after the update. The localization stack therefore favors bounded correction over either unstable free updates or premature frame rejection, which improves continuity when the robot passes through sparse or cluttered regions.

\subsection{Probabilistic Odometry Fusion and Multi-Floor Consistency}
\label{subsec:odom_multifloor}

Structural tracking is not used in isolation. The implementation augments it with a probabilistic odometry fusion module with adaptive weighting. After each scan correction, the system computes an ICP confidence score in $[0,1]$ from three terms: the ratio of valid correspondences, the consistency of the robust weights, and the mean point-to-line residual. This score determines how strongly the fused estimate should trust scan matching relative to odometric prediction:
\begin{equation}
    w_{\mathrm{ICP}} = 0.5 + 0.45 \cdot \mathrm{clamp}(s_{\mathrm{ICP}}, 0, 1), \qquad
    w_{\mathrm{odom}} = 1 - w_{\mathrm{ICP}},
\end{equation}
where $s_{\mathrm{ICP}}$ denotes the normalized ICP confidence score. The predicted pose is obtained from buffered wheel odometry through an AMCL-style motion model and interpolated to the LiDAR timestamp before fusion. Translation is combined by weighted averaging and orientation by quaternion slerp. Importantly, odometry acts as a predictive prior in the fusion layer; the scan-to-map correction remains the source of geometric alignment rather than being replaced by dead reckoning.

The same structure-first logic extends to multi-floor deployment. Area membership can be filtered by height compatibility before a planar inside-area test is accepted, and changes in the vertical offset of the \texttt{AGmap}/\texttt{map} relation can trigger floor-state updates. This prevents the estimator from accepting geometrically plausible but height-incompatible areas on adjacent levels, which is essential once planning, Rolling Window map serving, and localization all operate over the same multi-floor semantic map.

%% file: sections/7_sys_integration.tex
\section{System Integration and Engineering Contributions}
\label{sec:system_integration}

The preceding sections formalize the representation, architecture, and algorithms. This section addresses a different question: how these components are assembled into a deployable ROS2 stack. The emphasis is therefore on engineering integration rather than on architectural philosophy, covering the automated map-generation pipeline, the package-level software organization, the launch-level composition of the runtime system, and the configuration surfaces that support repeatable experiments.

\subsection{Automated Map Generation Pipeline}
\label{subsec:map_gen}

\begin{figure}[t]
    \centering
    \includegraphics[width=\linewidth]{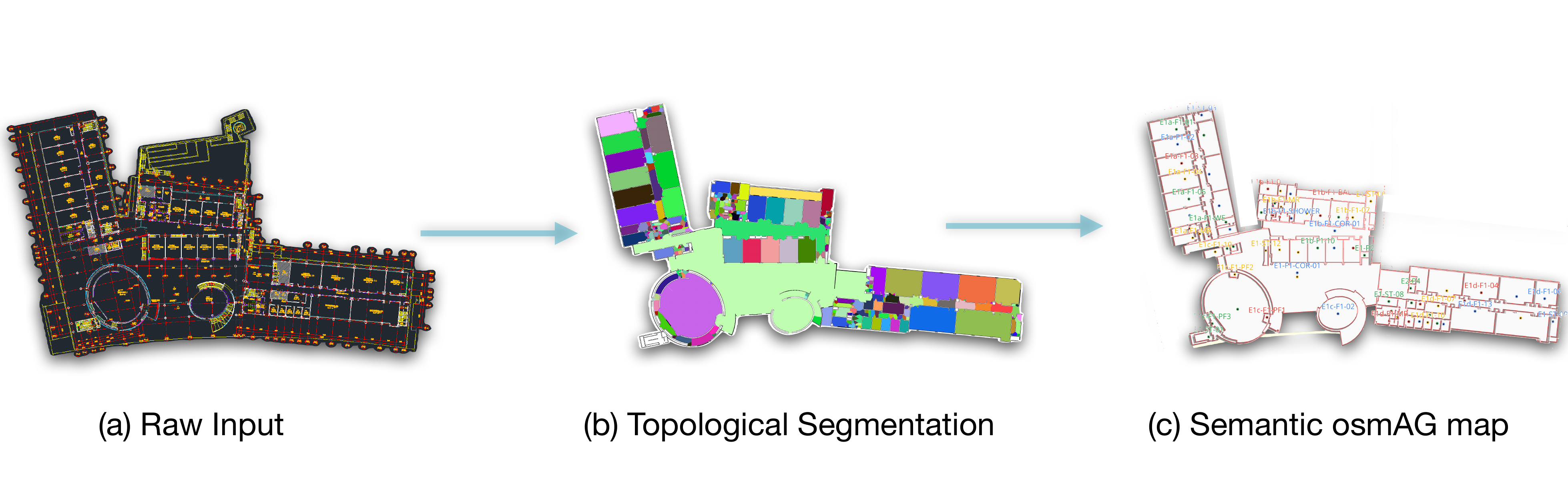}
    \caption{\textbf{Automated Map Generation Pipeline.} 
    (a) \textbf{Raw Input:} CAD file (`.dxf`) with unstructured geometry. 
    (b) \textbf{Topological Segmentation:} Automatic extraction of navigable Areas. 
    (c) \textbf{Semantic osmAG Output:} Final semantic vector map ready for navigation.}
    \label{fig:map_generation}
\end{figure}

To reduce the deployment burden associated with manual semantic map construction, the stack incorporates an automated CAD-to-osmAG pipeline~\citep{zhang2025generation}. As shown in Fig.~\ref{fig:map_generation}, the pipeline transforms standard architectural \texttt{.dxf} input through three stages: structural layer extraction, Voronoi-based topological segmentation into Areas, and automatic text association for semantic tagging. This toolchain addresses the ``cold start'' problem directly by producing a semantic-topological map before runtime navigation begins.

\subsection{Software Ecosystem and ROS2 Integration}
\label{subsec:software_ecosystem}

The runtime implementation is centered on two in-house ROS2 packages that are coupled to the standard Nav2 execution stack. Rather than collapsing the system into a single application binary, the software layout keeps environment serving, deliberative planning, and localization separable at the package and executable levels. In the deployed workflow, the principal runtime responsibilities are organized as follows.

\noindent\texttt{area\_graph\_data\_parser}\par
\begin{itemize}[leftmargin=1.5em,itemsep=2pt,topsep=3pt,parsep=0pt]
    \item \texttt{main}: parses osmAG, constructs the in-memory AreaGraph, and publishes structural map data for planning and localization.
    \item \texttt{local\_ag\_gridmap}: maintains the Rolling Window metric map and publishes \texttt{/local\_AGgridmap}.
    \item \texttt{osmag\_planner\_node}: executes global hierarchical passage-centric planning and segmented goal dispatch.
\end{itemize}

\noindent\texttt{localization\_using\_area\_graph}\par
\begin{itemize}[leftmargin=1.5em,itemsep=2pt,topsep=3pt,parsep=0pt]
    \item \texttt{cloud\_handler}: runs the core AGLoc pipeline, including rescue mode, point-to-line ICP tracking, corridor handling, and odometry fusion.
    \item \texttt{agloc\_localizer\_node}: adapts the AGLoc output to Nav2-compatible pose and TF interfaces via a lifecycle-managed adapter.
\end{itemize}

This decomposition is deliberate. The package \texttt{area\_graph\_data\_parser} groups semantic-topological map services and planner-side logic, whereas \texttt{localization\_using\_area\_graph} groups the AGLoc estimator and its ROS2 integration layer. The localization side is not only provided as standalone executables, but is also exported as reusable shared libraries and a ROS2 component, allowing the Nav2-facing adapter to remain composable without sacrificing conventional node-based bringup. In contrast to a monolithic navigation implementation, this structure keeps map processing, planning, and localization independently testable while still exposing standard ROS2 interfaces to the execution stack.

\subsection{Launch-Level Orchestration}

The runtime system is assembled through launch files rather than through a single hard-coded entry point. For the integrated planning-and-localization workflow used in end-to-end experiments, the central bringup script is \texttt{nav2\_planner\_loc.launch.py} in \texttt{area\_graph\_data\_parser}. This launch profile starts the semantic-topological environment services, the Rolling Window map server, the hierarchical planner, and the Nav2 bringup stack in a coordinated manner.

Launch-level orchestration matters for two reasons. It makes the runtime composition explicit, and it supports controlled experimentation without source modification. In addition to the integrated bringup above, the repository provides planning-oriented, visualization-oriented, and alternative Nav2 launch profiles within \texttt{area\_graph\_data\_parser}, as well as localization-only and AGLoc-to-Nav2 integration profiles within \texttt{localization\_using\_area\_graph}. This separation is useful for ablation, debugging, and subsystem validation because each profile exposes a different slice of the full stack while preserving the same ROS2 interfaces.

This distinction is operationally important. The integrated planning bringup explicitly starts \texttt{local\_ag\_gridmap}, whereas localization-focused launch paths do not necessarily do so. Reproducible experiments therefore depend not only on reporting the package names, but also on reporting the exact bringup profile and parameter set used for each evaluation.

\subsection{Configuration and Reproducibility}

Reproducibility in osmAG-Nav depends primarily on explicit package structure and parameterized launch configuration. Both primary packages install their launch files and runtime resources through standard CMake rules. In addition, the map-and-planner package installs the semantic-topological assets required for deployment, including \texttt{data}, \texttt{param}, and \texttt{maps}, while the localization package installs its \texttt{launch}, \texttt{config}, and \texttt{maps} resources. This organization makes the deployed stack relocatable within a ROS2 workspace rather than dependent on manually assembled scripts.

Runtime behavior is controlled through explicit parameter files rather than through hard-coded constants in launch wrappers. Planner-side behavior is configured jointly through the planner node parameters and the Nav2 parameter set in \texttt{area\_graph\_data\_parser}, while localization behavior is configured through the AGLoc parameter files in \texttt{localization\_using\_area\_graph}. Unlike a codebase in which major behaviors are buried in ad hoc scripts, the present implementation exposes architectural choices through ROS2-native configuration points: executable registration in CMake, package-scoped launch files, exported shared libraries, and package-scoped configuration resources. This organization makes it straightforward to reproduce integrated experiments, switch between bringup modes, and inspect the exact software composition used in each evaluation.

%% file: sections/8_experiment.tex
\section{Experiments and Evaluation}
\label{sec:experiments}

To rigorously validate the proposed \textit{osmAG-Nav} stack in authentic operating conditions, we conducted extensive real-world deployments at ShanghaiTech University. Departing from simulation-only validation, our experiments target the specific challenges of a live, operational campus environment. We evaluate the system across four critical dimensions: (1) data storage efficiency of the osmAG standard, (2) computational scalability of the hierarchical planner, (3) localization robustness in dynamic/degenerate environments, and (4) single-floor long-range navigation reliability.

\subsection{Experimental Setup}
\label{subsec:exp_setup}

\textbf{Environment:} Experiments were conducted at the School of Information Science and Technology (SIST), ShanghaiTech University. The dataset covers a large-scale multi-story teaching building with a total mapped area of approximately 11,025\,$m^2$ across multiple floors, featuring long featureless corridors (100\,m+), cluttered offices, glass-walled atriums, and elevators.

\textbf{Platform:} We utilized a custom-built Ackermann-steering mobile robot equipped with a Hesai PandarQT 64-line 3D LiDAR for perception and an Intel NUC (Core i7-1165G7) for onboard computation. The entire navigation stack runs on ROS2 Iron.

\subsection{Map Representation Efficiency}
\label{subsec:exp_storage}

We first evaluate the storage efficiency of the osmAG standard, a critical factor for scalability. We compared the memory footprint of our format against three standard representations for the same 11,025 $m^2$ environment: (1) a 2D Occupancy Grid (0.05m resolution) and (2) a downsampled 3D Point Cloud (0.2m voxel).

\begin{table}[htbp]
\centering
\caption{Comparison of Map Storage Efficiency (11,025\,$m^2$ Campus)}
\label{tab:storage_comparison}
\begin{tabular}{lcc}
\toprule
\textbf{Map Representation} & \textbf{Storage Size} & \textbf{Ratio (vs. osmAG)} \\
\midrule
2D Occupancy Grid (0.05m) & $\approx$ {48.5} MB & $34.6 \times$ \\
3D Point Cloud (0.1m voxel) & $\approx$ \text{1536}  MB & ${1097} \times$ \\
\textbf{osmAG (Vector Standard)} & \textbf{1.4 MB} & \textbf{1.0$\times$} \\
\bottomrule
\end{tabular}
\end{table}

As shown in Table~\ref{tab:storage_comparison}, osmAG achieves a reduction in storage size by two orders of magnitude compared to grid maps and three orders compared to point clouds. This lightweight nature ($<1$MB) enables the entire campus map to be transmitted instantly over low-bandwidth networks and loaded into RAM with negligible overhead, validating the ``System of Systems'' design philosophy where global state is decoupled from dense metric data.

\subsection{Hierarchical Planning Efficiency and Feasibility}
\label{subsec:exp_planning}

To systematically validate the computational scalability of the planning stack, we constructed a fixed benchmark set of 1,000 start-goal queries over the 11,025 $m^2$ campus map. The benchmark contains 100 short, 300 medium, 300 long, and 300 cross-floor tasks, spanning intra-room navigation, long corridor traversal, and multi-floor cross-building missions. Each identical query set was replayed under six different execution orders of Grid A*, Flat osmAG, and Hierarchical osmAG planning to suppress cache-order bias. The reported statistics are therefore based on robust case-level medians across these six runs rather than on a single execution order. We compare against two baselines: (1) standard \textbf{Nav2 Grid A*} on a high-resolution global costmap, and (2) a \textbf{Flat osmAG Planner} that searches the full multi-floor passage graph without LCA-based pruning.

\subsubsection{Computational Scalability and LCA Ablation}

\begin{figure}[htpb]
    \centering
    \includegraphics[width=0.95\linewidth]{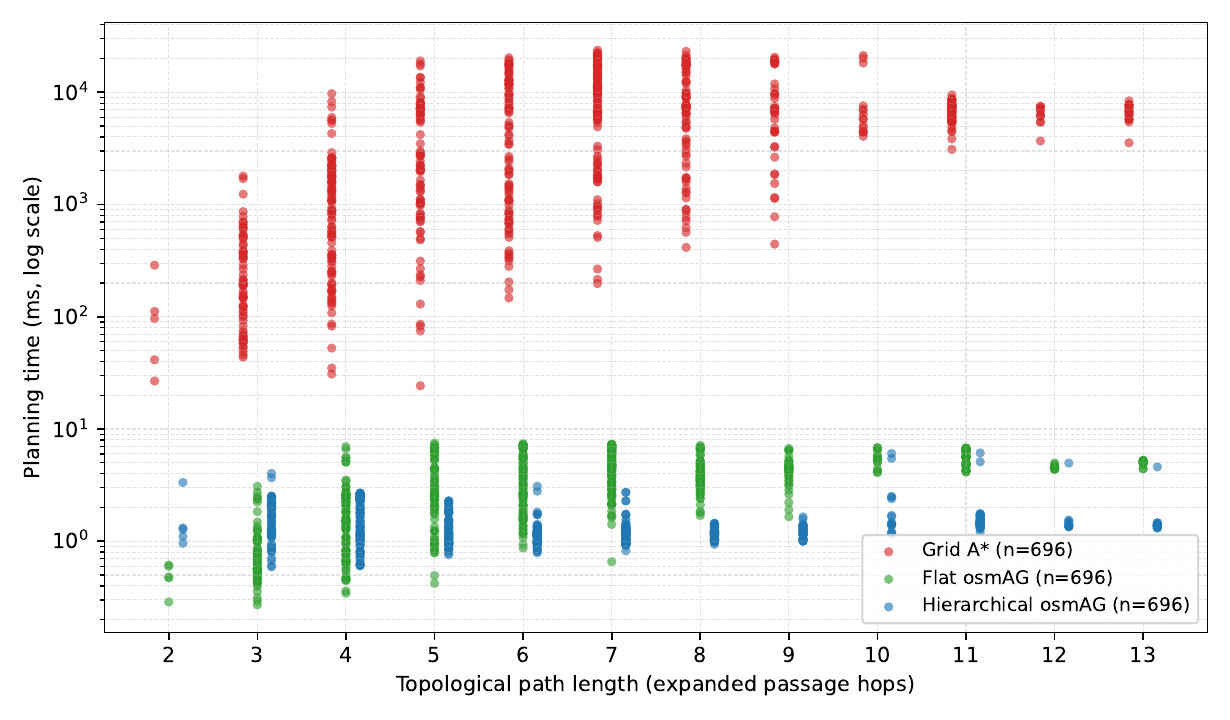}
    \caption{\textbf{Scalability Analysis of Global Planning.} Scatter plot of planning latency versus topological path length on the 696 same-floor queries for which all three planners are applicable. Each point is aggregated as the median over six execution orders, and the Y-axis is shown in log scale. Grid A* (red) exhibits rapid growth with route complexity, Flat osmAG (green) remains in the low-millisecond regime, and Hierarchical osmAG (blue) overtakes the flat baseline at roughly four hops before maintaining a nearly flat latency profile thereafter.}
    \label{fig:planning_scatter}
\end{figure}

Because a 2D global grid planner is not a meaningful comparator for cross-floor routing, the three-way comparison in Table~\ref{tab:planning_stats} and Fig.~\ref{fig:planning_scatter} is restricted to the 696 same-floor queries for which Grid A*, Flat osmAG, and Hierarchical osmAG are all applicable. On this common comparable subset, the limitation of monolithic metric planning is immediately visible: the mean planning latency of Grid A* rises from $204.3$ ms on short queries to $2879.3$ ms on medium queries and $10641.0$ ms on long queries, while the mean number of closed grid states grows from $1.96\times 10^4$ to $3.68\times 10^5$. Such growth is incompatible with responsive long-horizon replanning in a campus-scale environment.

Replacing the global grid with a sparse passage graph already collapses the search space substantially. The Flat osmAG baseline reduces the mean end-to-end planning time to $0.757$ ms, $2.939$ ms, and $5.308$ ms in the short, medium, and long buckets, respectively, while preserving low path-length overhead relative to Grid A* ($4.29\%$, $2.66\%$, and $2.07\%$). The proposed \textbf{hierarchical osmAG planner} preserves the same path-quality overhead but reduces the explored search space further, down to only $19$, $24$, and $21$ mean closed passage states in the three buckets. The end-to-end latency picture is intentionally more nuanced than a pure search-kernel comparison: the hierarchy introduces a small fixed \textit{attach}$\rightarrow$\textit{lift}$\rightarrow$\textit{expand} overhead, so Flat osmAG remains slightly faster on the shortest routes ($0.757$ ms versus $1.308$ ms). Once query complexity reaches the medium and long regimes, however, LCA-based pruning becomes decisively beneficial, reducing the mean planning time to $1.443$ ms and $1.361$ ms, corresponding to $1995\times$ and $7816\times$ speedups over Grid A*.

Figure~\ref{fig:planning_scatter} makes the same trend visible at the individual-query level. The hierarchical planner overtakes the flat baseline at roughly four topological hops and then maintains an almost distance-insensitive latency profile, whereas Grid A* grows rapidly with route complexity. This is the intended role of the hierarchy: not a universal constant-factor win for every tiny query, but a reliable mechanism for preventing search explosion on long building-scale routes. Although excluded from the three-way same-floor figure, the 297 successful cross-floor graph queries show the same tendency: the hierarchical planner averages $2.02$ ms, compared with $4.76$ ms for the flat planner, while retaining native multi-floor routing through the passage graph.

\begin{table}[htpb]
\centering
\caption{Quantitative benchmark of global planning efficiency on the common comparable same-floor subset. The original benchmark contains 1,000 fixed queries, but the three-way comparison is restricted to the 696 same-floor queries for which Grid A*, Flat osmAG, and Hierarchical osmAG are all applicable. Reported times are end-to-end planning latencies, aggregated as case-level medians across six execution orders and then averaged within each task bucket. Overhead denotes the mean path-length increase relative to Grid A*.}
\label{tab:planning_stats}
\tiny
\setlength{\tabcolsep}{1.5pt}
\begin{tabularx}{0.96\columnwidth}{@{}>{\raggedright\arraybackslash}p{0.22\columnwidth}>{\raggedright\arraybackslash}p{0.16\columnwidth}>{\centering\arraybackslash}p{0.11\columnwidth}>{\centering\arraybackslash}p{0.10\columnwidth}>{\centering\arraybackslash}p{0.14\columnwidth}>{\centering\arraybackslash}X@{}}
\toprule
\begin{tabular}[c]{@{}c@{}}\textbf{Task}\\\textbf{Complexity}\end{tabular} &
\textbf{Planner} &
\begin{tabular}[c]{@{}c@{}}\textbf{Time}\\\textbf{(ms)}\end{tabular} &
\textbf{Gain} &
\textbf{Overhead} &
\textbf{States$^\ast$} \\
\midrule
\multirow{3}{*}{\begin{tabular}[c]{@{}l@{}}\textbf{Short}\\ ($<50$m, 1--3 Hops)\end{tabular}} 
& Nav2 Grid A* & 204.3 & Baseline & Optimal & 19,639 \\
& Flat osmAG & 0.757 & 270$\times$ & +4.29\% & 32 \\
& \begin{tabular}[c]{@{}l@{}}\textbf{Hierarchical}\\\textbf{osmAG (Ours)}\end{tabular} & 1.308 & 156$\times$ & +4.29\% & 19 \\
\midrule
\multirow{3}{*}{\begin{tabular}[c]{@{}l@{}}\textbf{Medium}\\ (50--150m, 4--6 Hops)\end{tabular}} 
& Nav2 Grid A* & 2879.3 & Baseline & Optimal & 140,320 \\
& Flat osmAG & 2.939 & 980$\times$ & +2.66\% & 155 \\
& \begin{tabular}[c]{@{}l@{}}\textbf{Hierarchical}\\\textbf{osmAG (Ours)}\end{tabular} & 1.443 & 1,995$\times$ & +2.66\% & 24 \\
\midrule
\multirow{3}{*}{\begin{tabular}[c]{@{}l@{}}\textbf{Long}\\ ($>150$m, $>$6 Hops)\end{tabular}} 
& Nav2 Grid A* & 10641.0 & Baseline & Optimal & 367,581 \\
& Flat osmAG & 5.308 & 2,005$\times$ & +2.07\% & 292 \\
& \begin{tabular}[c]{@{}l@{}}\textbf{Hierarchical}\\\textbf{osmAG (Ours)}\end{tabular} & 1.361 & 7,816$\times$ & +2.07\% & 21 \\
\bottomrule
\end{tabularx}

\vspace{1ex}
\parbox{0.94\columnwidth}{\tiny\raggedright $^\ast$\textit{States denote the mean number of closed states in each planner's native search space}: Grid A* expands metric grid cells, whereas osmAG planners expand semantic passages. The Flat-versus-Hierarchical comparison therefore isolates the effect of LCA pruning rather than metric discretization.}
\end{table}

\subsubsection{Static-Cache Ablation}

The previous benchmark establishes that hierarchical planning scales well as route complexity grows. We next examine why this is the case. To isolate the contribution of the hierarchical static caches, we ran a paired ablation on 20 fixed queries (5 short, 5 medium, 5 long, and 5 cross-floor), with five cached/uncached trials per query. In this experiment, Grid A* was intentionally disabled so that the comparison reflects only the internal behavior of \texttt{PlanMultiLayerPath()}. The cached setting reuses the prebuilt area-compact, parent-lift, and connectivity caches, whereas the uncached setting clears these structures before each query and forces them to be rebuilt online.

\begin{table}[htbp]
\centering
\small
\setlength{\tabcolsep}{5pt}
\caption{\textbf{Ablation of hierarchical static caches.} Results are aggregated over 20 fixed benchmark cases with 5 cached/uncached paired trials each. ``A* only'' denotes the accumulated internal A* search time, whereas ``method wall'' is the end-to-end hierarchical query latency.}
\label{tab:hier_cache_ablation}
\begin{tabular}{lcccc}
\toprule
\textbf{Setting} & \textbf{Wall (ms)} & \textbf{A* (ms)} & \textbf{Rebuild (ms)} & \textbf{Equal Path} \\
\midrule
Cached & 1.827 & 0.336 & -- & -- \\
Uncached & 106.011 & 0.347 & 104.197 & 100/100 \\
\bottomrule
\end{tabular}
\end{table}

Table~\ref{tab:hier_cache_ablation} shows that removing the static caches increases the mean hierarchical query latency from $1.827$ ms to $106.011$ ms, corresponding to a $68.49\times$ slowdown. More importantly, the mean extra latency ($104.183$ ms) is numerically explained almost entirely by the mean cache-rebuild cost ($104.197$ ms), while the internal A* time changes only from $0.336$ ms to $0.347$ ms. The same pattern appears in every task category: the mean slowdown is $93.20\times$ for short queries, $65.02\times$ for medium queries, $59.66\times$ for long queries, and $56.09\times$ for cross-floor queries. Because all 100 cached/uncached pairs produce identical paths and path costs, this ablation indicates that the benefit of the hierarchical implementation comes primarily from moving compact-graph preparation out of the online query path rather than from altering route quality.

\begin{figure*}[htpb]
    \centering
    \includegraphics[width=0.98\textwidth]{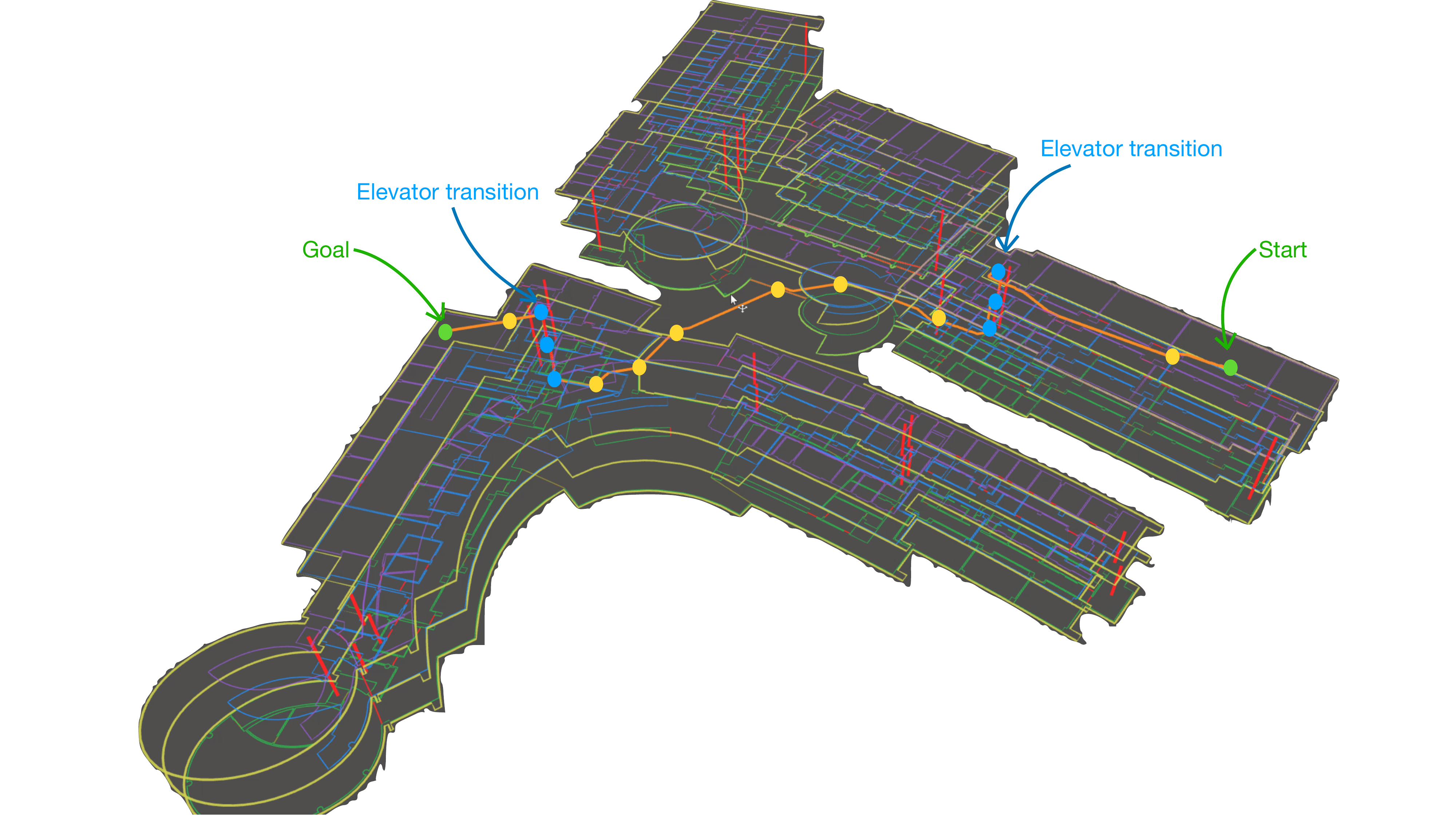}
    \caption{\textbf{RViz Visualization of an Actual Hierarchical Planning Result.} A real global route produced by the deployed Hierarchical osmAG planner and visualized in RViz for a multi-building, multi-floor mission. The route connects the start and goal across buildings, traverses the outdoor connector, and uses elevator transitions as explicit vertical passages. The highlighted passage nodes reveal the sparse topological decision sequence underlying the route, showing that cross-floor and cross-building navigation are handled within one unified hierarchical planning process rather than by separate floor-switching logic.}
    \label{fig:planning_viz}
\end{figure*}

Where Table~\ref{tab:planning_stats}, Fig.~\ref{fig:planning_scatter}, and Table~\ref{tab:hier_cache_ablation} quantify scalability and its implementation source, Fig.~\ref{fig:planning_viz} grounds these results in one concrete deployed query. Specifically, it shows an actual campus-scale planning result produced by the deployed Hierarchical osmAG planner and visualized in RViz. Rather than comparing dense path traces qualitatively, the figure is intended to explain how a single long-horizon query is realized as a sparse topological decision sequence that spans buildings, traverses the outdoor connector, and incorporates elevator transitions directly into one unified route.

\subsection{Localization Robustness Evaluation}
\label{subsec:exp_localization}

We evaluated the robustness of \textbf{AGLoc++}, the localization module used in \textit{osmAG-Nav}, against two baselines: the original \textbf{AGLoc} framework~\citep{xie2023robust} and \textbf{AMCL} as the standard grid-map localization baseline in Nav2. The tracking benchmark covers both cluttered laboratory sequences and geometrically degenerate corridor sequences. The corridor cases are particularly challenging because they combine weak longitudinal structure with frequent local ambiguity during long traversals:
\begin{itemize}
    \item \textbf{Dynamic Clutter:} Pedestrians frequently occluded the robot's view.
    \item \textbf{Geometric Degeneracy:} A 40m section featured smooth walls with no distinctive geometric features.
\end{itemize}

\begin{figure}[htbp]
    \centering
    \includegraphics[width=0.78\columnwidth]{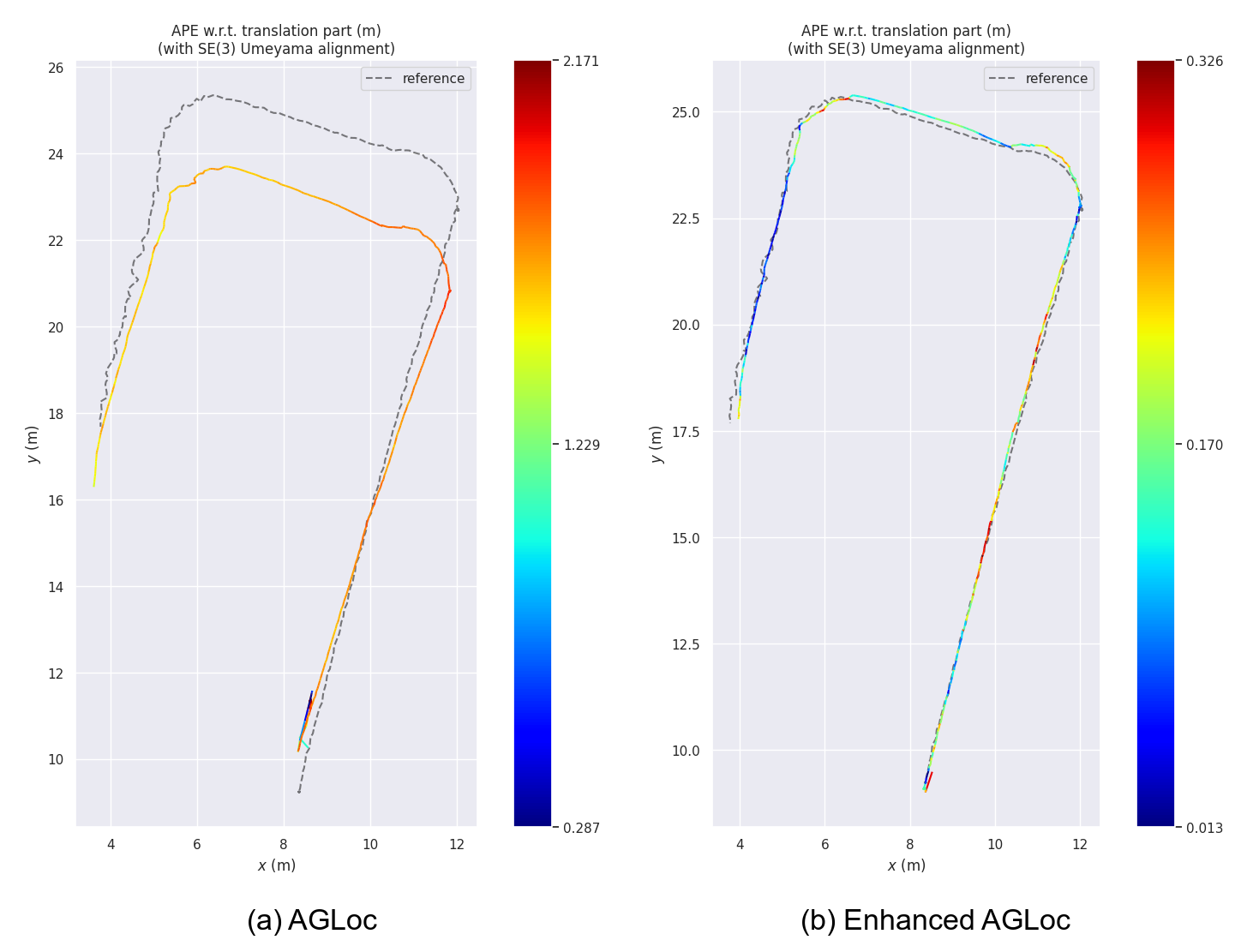}
    \caption{\textbf{Localization Comparison in a Geometrically Degenerate Corridor.} (a) Baseline AGLoc exhibits substantial drift relative to the reference trajectory on the corridor sequence \texttt{TK-Corr-01}, as shown by the large-error APE heatmap. (b) AGLoc++, as used in \textit{osmAG-Nav}, remains tightly aligned with the reference and markedly suppresses longitudinal slip. These results indicate that corridor-aware constraint handling and adaptive odometry fusion materially improve localization robustness in degenerate indoor scenes.}
    \label{fig:loc_comparison}
\end{figure}

\textbf{Metric:} Absolute Trajectory Error (ATE) RMSE relative to a high-precision ground truth. Table~\ref{tab:loc_tracking_summary} reports the aggregate tracking accuracy and per-frame runtime, while Fig.~\ref{fig:loc_comparison} visualizes a representative corridor sequence.

\begin{table}[htbp]
\centering
\small
\setlength{\tabcolsep}{4pt}
\caption{\textbf{Compact Localization Tracking Summary.} ATE RMSE is averaged over two laboratory sequences (\texttt{TK-Mars}, \texttt{TK-Door}) and three corridor sequences (\texttt{TK-Corr-01/02/03}).}
\label{tab:loc_tracking_summary}
\begin{tabular}{lccc}
\toprule
\textbf{Method} & \begin{tabular}[c]{@{}c@{}}\textbf{Frame Time}\\\textbf{(ms)}\end{tabular} & \begin{tabular}[c]{@{}c@{}}\textbf{Lab ATE}\\\textbf{(m)}\end{tabular} & \begin{tabular}[c]{@{}c@{}}\textbf{Corridor ATE}\\\textbf{(m)}\end{tabular} \\
\midrule
AMCL & 17.2 & 0.56 & 1.52 \\
AGLoc~\citep{xie2023robust} & 12.5 & 0.08 & 1.26 \\
\textbf{AGLoc++ (ours)} & \textbf{15.6} & \textbf{0.07} & \textbf{0.36} \\
\bottomrule
\end{tabular}
\end{table}

Table~\ref{tab:loc_tracking_summary} shows that AGLoc++ preserves the already strong laboratory accuracy of the original AGLoc baseline while materially improving robustness in corridors, reducing the corridor-average ATE RMSE from $1.26$m to $0.36$m. It also substantially outperforms AMCL in both environment groups, especially in the corridor regime where grid-map localization suffers from degeneracy and scene variation. The representative case in Fig.~\ref{fig:loc_comparison} illustrates the same effect qualitatively: on \texttt{TK-Corr-01}, the baseline AGLoc trajectory deviates strongly from the reference, whereas AGLoc++ remains closely aligned with it, reducing the peak absolute position error from approximately $2.17$m to $0.33$m. Taken together, the aggregate table and the per-sequence visualization support the claim that corridor-aware constraint selection and adaptive odometry fusion materially strengthen localization robustness in long-horizon deployment.

\subsection{Long-Range Navigation Reliability}
\label{subsec:exp_nav_reliability}

To stress-test the integrated stack at mission scale, we conducted a \textbf{Long-Range Single-Floor Mission Experiment} in which the robot autonomously navigates a 235\,m route within a single floor of the SIST building, covering a floor area of approximately 2,300\,$m^2$. This mission exercises the core capabilities emphasized throughout the paper: long-horizon hierarchical routing across multiple sectors and corridors, Rolling Window map serving, and segmented goal dispatch through Nav2. The evaluation protocol consists of 10 consecutive end-to-end trials so that both route feasibility and long-range execution reliability can be assessed under repeated deployment.

\textbf{Mission-Level Results.}
Over the 10 trials, osmAG-Nav achieved a success rate of 9\,/\,10, with an average mission duration of 7\,min\,37\,sec ($\pm$\,8\,sec). Each successful trial required the planner to coordinate hierarchical passage-level routing across multiple sectors of the floor, dispatch intermediate goals through long corridors and passage transitions, and maintain continuous localization against permanent structure---all without manual intervention. The single failure occurred when a crowd of bystanders congregated in front of the elevator lobby along the planned route, inflating the local costmap costs to the point where the Nav2 controller server could no longer produce a feasible trajectory; after exhausting its configured recovery behaviors (e.g., wait, backup, spin), the Nav2 Behavior Tree returned \texttt{FAILURE}, causing the \texttt{NavigateToPose} action to abort the mission.

\textbf{Execution Continuity Analysis.}
As a complementary execution-side assessment, we examined a representative $49.26$\,s deployment segment to determine whether Rolling Window map refresh and segmented goal dispatch introduce visible motion interruption. The log contains 937 odometry-derived speed samples at an average rate of approximately $19$\,Hz, together with two \texttt{map\_switch} events and two internal Nav2 goal-switch events. The instantaneous planar-speed perturbation remained small for all four switches: the two map updates produced first-frame changes of $-0.0046$\,m/s and $-0.0027$\,m/s, while the two goal switches produced $-0.0039$\,m/s and $+0.0091$\,m/s. Over the more reliable $0.1$--$0.2$\,s post-event windows, no systematic speed collapse was observed. Because one goal switch and one map switch occur only $0.328$\,s apart, the longer $0.5$--$1.0$\,s windows are interpreted with care. Overall, the observed event-level response is consistent with bounded local-map updates and segmented handoff being executed without severe stop-and-go behavior.

Taken together, the mission-level success rate and the event-level continuity analysis confirm that \textit{osmAG-Nav} can reliably execute long-range missions under repeated deployment, and that the Rolling Window and Segmented Execution mechanisms operate without introducing significant motion degradation.

%% file: sections/9_conclusion.tex
\section{Discussion and Conclusion}
\label{sec:conclusion}

\subsection{Summary}
\label{subsec:summary}

In this paper, we presented \textit{osmAG-Nav}, a complete, open-source navigation stack for ROS2 that replaces the dense occupancy-grid world model with a sparse, hierarchical semantic-topometric representation grounded in the \textit{osmAG} standard. By decoupling global topological reasoning from local metric execution, the system addresses the scalability, multi-floor reasoning, and long-term robustness bottlenecks inherent in conventional grid-centric navigation frameworks.

The principal contributions and findings are as follows. We formalized the \textbf{osmAG standard} as a general-purpose spatial representation for robotics and embodied AI, and built a \textbf{decoupled ``System of Systems'' architecture} that preserves the mature local planning and control capabilities of Nav2 while replacing the global world model with a semantic-topological stack. The \textbf{Rolling Window} mechanism keeps the local costmap memory footprint bounded independently of the total mapped area, bridging global vector topology with local metric execution. The \textbf{Hierarchical osmAG Planning Engine}, anchored by an LCA-based decomposition on a \textit{passage-centric graph} with \textit{segmented execution}, suppresses search explosion on building-scale routes: in the comparable same-floor benchmark subset, Grid A* grows from $204.3$\,ms to $10641.0$\,ms from the short to the long bucket, whereas the hierarchical planner remains in the low-millisecond regime and delivers $1995\times$ and $7816\times$ speedups in the medium and long buckets. The \textbf{structure-aware localization framework}, building upon AGLoc, strengthens tracking stability in dynamic and geometrically degenerate environments through direction-aware constraint selection, adaptive odometry fusion, and corridor-aware degeneracy handling, reducing corridor-average ATE~RMSE from $1.26$\,m to $0.36$\,m. Extensive \textbf{real-world experiments} on a multi-story campus ($>$11,025\,m$^2$) validate these capabilities: offline benchmarks confirm planning scalability and localization robustness, while a single-floor long-range robot mission (235\,m) demonstrates that the integrated stack sustains reliable autonomous operation over extended distances where conventional grid-based stacks degrade.

\subsection{Limitations}
\label{subsec:limitations}

Despite the system's demonstrated robustness, we acknowledge specific limitations inherent to our topology-first design philosophy:

\begin{itemize}
    \item \textbf{Dependency on Structured Topology:} Our hierarchical planner relies on the assumption that the environment can be cleanly segmented into ``Areas'' connected by ``Passages.'' In highly unstructured environments (e.g., open-plan warehouses with virtual zones or unstructured outdoor plazas), the automatic segmentation may produce ambiguous topologies, potentially leading to suboptimal routing compared to free-space metric planners.
    
    \item \textbf{Static Topological Assumption:} While our localization module effectively filters dynamic clutter (pedestrians, furniture), the underlying topological graph is assumed to be structurally static. If a permanent architectural change occurs—such as a new door being installed or a corridor being partitioned—the system currently lacks an online mechanism to modify the graph topology. This requires an offline update of the osmAG database to restore connectivity.
    
\end{itemize}

\subsection{Future Work}
\label{subsec:future_work}

Building upon this foundation, future research will focus on extending the adaptability and intelligence of the system:

\begin{itemize}
    \item \textbf{Lifelong Topological Maintenance:} To address the static topology limitation, we intend to develop a ``Lifelong Mapping'' module. By detecting persistent discrepancies between the vector map and sensor data, the system could propose topological updates (e.g., ``Passage Blocked'' or ``New Connection Detected'') and autonomously update the osmAG database during operation.
    
    \item \textbf{Semantic-Aware Cost Learning:} The present edge weights are physically-aware but remain raster-derived geometric cost proxies. We plan to explore learning-based cost functions that incorporate rich semantic context—such as penalizing ``Quiet Zones'' or prioritizing ``Public Corridors'' during peak hours—to fully exploit the semantic capabilities of the osmAG format.
    
    \item \textbf{Heterogeneous Fleet Coordination:} The hierarchical nature of our planner is well-suited for multi-robot systems. We aim to extend the stack to support fleet coordination, where the central topological graph serves as a shared resource for reserving passages (critical resources like elevators) to prevent congestion among multiple robots.

    \item \textbf{Cross-Floor Real-Robot Deployment:} The osmAG standard natively encodes multi-floor hierarchy and vertical topology (elevators, stairs), and the hierarchical planner already fully supports cross-floor route planning as demonstrated in Fig.~\ref{fig:planning_viz}. However, closing the loop on physical cross-floor execution requires integrating several additional engineering modules: a Behavior Tree orchestrating elevator approach, door-open/close interaction, and cabin entry/exit within the Nav2 framework; robust floor-level introspection so that the robot can determine which floor it currently occupies, for example via differential barometric altimetry~\citep{zhang2025differential}; and a reliable transition handoff that re-anchors localization after the vertical transfer. These integration challenges do not affect the core algorithmic contributions of this work---hierarchical planning, Rolling Window rasterization, and structure-aware localization---but they constitute a substantial systems-engineering effort that we plan to address as future work.
\end{itemize}

%% file: sections/10_declarations.tex
\section*{Declarations}

\begin{itemize}
\item \textbf{Funding}
This work was supported by the School of Information Science and Technology at ShanghaiTech University.

\item \textbf{Conflict of interest}
The authors have no competing interests to declare that are relevant to the content of this article.

\item \textbf{Code availability}
The complete source code for the osmAG-Nav stack, including the planning and localization packages, as well as full documentation and Docker containers, will be made publicly available to facilitate reproducibility at the time of publication.

\item \textbf{Author contribution}
\textbf{Yongqi Zhang} and \textbf{Jiajie Zhang} contributed equally to this work. They developed the core navigation stack, implemented the hierarchical planner, and conducted the experiments. \textbf{Chengqian Li} developed the flat osmAG planner module. \textbf{Fujing Xie} developed the localization module and contributed to the system integration. \textbf{Soeren Schwertfeger} supervised the project, provided theoretical guidance, and revised the manuscript.
\end{itemize}

%% file: sections/appendix_osmag_reference.tex
\section{Detailed osmAG Tag Reference}
\label{sec:appendix_osmag_reference}

This appendix collects the detailed tag-level reference that complements the normative summary in Section~\ref{sec:osmag}. The appendix is intentionally broader than the main-text core schema: it includes the canonical core tags and the sanctioned extension profiles that are most useful as a compact reference during reading.

\begin{table*}[p]
\caption{Detailed reference for canonical core and core-supporting osmAG fields.}
\label{tab:osmag_detailed_core_reference}
\centering
\scriptsize
\setlength{\tabcolsep}{3pt}
\renewcommand{\arraystretch}{1.0}
\begin{tabularx}{\textwidth}{@{}>{\raggedright\arraybackslash}p{0.23\textwidth}>{\raggedright\arraybackslash}p{0.14\textwidth}>{\raggedright\arraybackslash}p{0.18\textwidth}>{\raggedright\arraybackslash}p{0.14\textwidth}X@{}}
\toprule
\textbf{Key or form} & \textbf{Namespace / alias} & \textbf{Scope} & \textbf{Status} & \textbf{Meaning and usage} \\
\midrule
\texttt{name=root} & OSM node tag & Core anchor & Core & Declares the unique root node that anchors the local Cartesian frame to geodetic coordinates. \\
\texttt{name} (area) & OSM way tag & Core identifier & Core & Globally unique canonical identifier of an area; the reference target for \texttt{osmAG:parent}. \\
\texttt{name} (passage) & OSM way tag & Core identifier & Core & Globally unique canonical identifier of a passage. \\
\texttt{osmAG:type=area} & \texttt{osmAG:*} & Core topology & Core & Declares a way as an osmAG area. \\
\texttt{osmAG:type=passage} & \texttt{osmAG:*} & Core topology & Core & Declares a way as an osmAG passage. \\
\texttt{osmAG:areaType} & \texttt{osmAG:*} & Core semantics & Core & Canonical area class field; normative values include \texttt{room}, \texttt{corridor}, \texttt{structure}, \texttt{elevator}, and \texttt{stairs}. \\
\texttt{osmAG:parent} & \texttt{osmAG:*} & Core hierarchy & Core & Parent-area reference using the canonical semantic \texttt{name}. \\
\texttt{osmAG:from}, \texttt{osmAG:to} & \texttt{osmAG:*} & Core topology & Core & Unordered incident-area references of a passage, encoded with canonical semantic names. \\
\texttt{level} & OSM-compatible tag & Core vertical indexing & Core & Floor label used for multi-floor structure and vertical reasoning. \\
\texttt{height} & OSM-compatible tag & Core elevation & Core & Elevation above the ground-floor datum; optional in general and used by some vertical links. \\
\texttt{indoor} & OSM-compatible tag & Interoperability & Core-supporting & OSM interoperability tag commonly used by authoring and visualization tools; not the canonical osmAG semantic class field. \\
\botrule
\end{tabularx}
\end{table*}

\begin{table*}[p]
\caption{Detailed reference for sanctioned osmAG extension profiles.}
\label{tab:osmag_detailed_extension_reference}
\centering
\scriptsize
\setlength{\tabcolsep}{3pt}
\renewcommand{\arraystretch}{1.0}
\begin{tabularx}{\textwidth}{@{}>{\raggedright\arraybackslash}p{0.23\textwidth}>{\raggedright\arraybackslash}p{0.16\textwidth}>{\raggedright\arraybackslash}p{0.15\textwidth}>{\raggedright\arraybackslash}p{0.12\textwidth}X@{}}
\toprule
\textbf{Key or form} & \textbf{Namespace / alias} & \textbf{Scope} & \textbf{Status} & \textbf{Meaning and usage} \\
\midrule
\texttt{osmAG:area\_usage} & \texttt{osmAG:*} & Navigation semantics & Extension & Optional functional role of an area, for example lobby, classroom, office, or lab. \\
\texttt{osmAG:room\_number} & \texttt{osmAG:*} & Navigation semantics & Extension & Human-facing room number distinct from the globally unique canonical \texttt{name}. \\
\texttt{osmAG:occupied\_by} & \texttt{osmAG:*} & Navigation semantics & Extension & Stable ownership, tenancy, or organizational occupancy metadata. \\
\texttt{osmAG:passage\_type} & \texttt{osmAG:*} & Navigation semantics & Extension & Optional passage semantics such as automatic, handle-operated, or sliding. \\
\texttt{osmAG:degree} & \texttt{osmAG:*} & Simulation metadata & Extension & Optional articulation metadata for simulation; not required by the map standard itself. \\
\shortstack[l]{\texttt{semantic\_osmAG:}\\\texttt{object\_name}} & \shortstack[l]{\texttt{semantic\_}\\\texttt{osmAG:*}} & Embodied-AI semantics & Extension & Named object instance attached to an OSM node with explicit coordinates. \\
\shortstack[l]{\texttt{semantic\_osmAG:}\\\texttt{observed\_objects}} & \shortstack[l]{\texttt{semantic\_}\\\texttt{osmAG:*}} & Embodied-AI semantics & Extension & Viewpoint-centric open-vocabulary observation summary attached to a node. \\
\shortstack[l]{\texttt{semantic\_osmAG:}\\\texttt{area\_description}} & \shortstack[l]{\texttt{semantic\_}\\\texttt{osmAG:*}} & Embodied-AI semantics & Extension & Natural-language room or area description attached to an area way. \\
\botrule
\end{tabularx}
\end{table*}